\documentstyle[11pt]{article}

\parskip=\medskipamount

\begin{document}
\thispagestyle{empty}

\begin{center}
{\Large \bf 
  Report of the Study Group on\\[1ex] Assessment and Evaluation}\\[8mm]
{\large April 1995}\\[8mm]

Edited by\\[2mm]
\begin{tabular}{ccc}
Richard Crouch   & \hspace*{4em} & Robert Gaizauskas\\
SRI International          & & Dept. Computer Science    \\
Cambridge Research Centre  & & University of Sheffield   \\
23 Millers Yard, Mill Lane & & Regent Court, 211 Portobello St \\
Cambridge CB2 1RQ, UK      & & Sheffield S1 4DP, UK       \\
{\tt rc@cam.sri.com}       & & {\tt R.Gaizauskas@dcs.shef.ac.uk} 
\end{tabular}

\bigskip

\begin{tabular}{c}
 Klaus Netter\\
 DFKI, Saarbr\"ucken\\
 Stuhlsatzenhausweg 3\\
 D-66123 Saarbr\"ucken\\
 Germany\\
 {\tt netter@dfki.uni-sb.de}
\end{tabular}

\bigskip
\bigskip

With additional contributions from

\bigskip

\begin{tabular}{ccc}
Geert Adriaens  & Karen Sparck Jones   & Herman Steeneken\\
Siemens Nixdorf & Cambridge University & TNO Inst. for Human Factors
\end{tabular}
\end{center}

\vspace*{10mm}
\rule{\columnwidth}{.02in}

This document was prepared at the request of the European Commission DG XIII
by an ad hoc study group.  It is a draft interim report now being made
available in the form in which it was submitted to the Commission.  It is
not an official European Commission document, and does not reflect the
policy of the European Commission.

\rule{\columnwidth}{.02in}\\

\newpage

\thispagestyle{empty}
\begin{center} {\Large Abstract} \end{center}

\bigskip
This is an interim report discussing possible guidelines for the assessment and
evaluation of projects developing speech and language systems.  It was prepared
at the request of the European Commission DG XIII by an ad hoc study group,
and is now being made available in the form in which it was submitted to the
Commission.  However, the report is not an official European Commission
document, and does not reflect European Commission policy,  official or
otherwise.   After a
discussion of terminology, the report focusses on combining user-centred and 
technology-centred assessment, and on how meaningful comparisons can be made
of a variety of systems performing different tasks for different domains.
The report outlines the kind of infra-structure that might be required to
support comparative assessment and evaluation of heterogenous projects, and
also the results of a questionnaire concerning different approaches to 
evaluation.

\newpage
\setcounter{page}{1}
\tableofcontents
\newpage

\section{Introduction}

This is an interim report of the study group on evaluation set up to
work out guidelines for assessment and validation of projects in the
Language Engineering sector (LE) of the TELEMATICS Fourth Framework
Programme (FP-4).

\subsection{The Study Group on Assessment}

In October 1994 a number of experts were invited by the CEC to a meeting
whose purpose was to initiate a Study Group on Assessment. The task of
this group is to work out guidelines and specifications for assessment and
validation of LE projects in the Fourth Framework Programme (FP-4),
specifically those responding to the first and second call for proposals
in December 1994 and September 1995 respectively.

The SG has its organisational framework within EAGLES. It is chaired by
EC staff and supported by three to four part-time editors or
rapporteurs, whose task is to coordinate the actions of the study group,
to support the composition of trigger papers on various topics, and to
work out the proposals for actions within FP-4. The study group will
have completed its task by end of June 1995 and cease to exist
thereafter.

\subsection{Mandate and Scope of the Study Group}

Assessment in the FP-4 covers a broad range of activities, ranging from
the TELEMATICS Programme level, through the sector level to the level of
individual projects.

At programme level, assessment is meant to address issues such as the
following:

\begin{itemize}

\item assessing value for money

\item assessing the programme's contribution to competitiveness of
      Community industry

\item identifying ways and means to maximise effectiveness of RTD policy

\end{itemize}

Preparatory work is under way to provide the basic assessment frameworks
required to identify the main assessment and evaluation issues for
Programme Support Actions (in particular SU~1.3 of the TELEMATICS
Application Work Plan), and to determine expertise requirements and
likely assessment issues needed. 

The SU~1.3. assessment task has connections with SU~1.1. Strategic
Marked and Innovation Watch, as well as with sector-based tasks
concerned with assessment and evaluation.

While recognising the importance of the above and the need for a
coherent approach to all assessment issues, the mandate and scope of the
study group is limited to issues at the level of the LE sector.
Specifically it concentrates on the preparation of a proposal for
setting up infrastructure and guidelines for technology assessment and
performance analysis, including a comparative evaluation where possible,
taking into account the broader context of project assessment, user
validation and field testing and the user driven paradigm of the
TELEMATICS APPLICATIONS in general and the LE sector in particular.

The activities of the study group are thus limited to assessment issues
at the level of individual projects and project clusters in the LE
sector, as outlined by the following:

\begin{itemize}

\item 
Within individual projects:\\
Projects responding to calls for proposals have to take assessment issues
into account from the very beginning.  Proposals must outline methods for
verifying and validating project results,  and dedicate a certain amount
of effort to evaluation effort during project cycles.

Assessment and evaluation is accordingly conducted primarily by the project 
consortium members themselves, but additionally by peer reviewers and EC
consultants who will monitor work on an {\em ad hoc} basis.

\item
Within project clusters:\\
Project clusters may be formed on the basis of related objectives and
applications (e.g. technical authoring, information access, information
management), or on the basis of common user groups.

Shared tasks within project clusters could include, amongst others, market
research and user requirement definitions, the establishment of user forums,
data gathering, the testing and assessment of generic technologies,
and the development of user validation guidelines.  Appropriate
organisational structures to support these shared tasks need to be
identified.

\end{itemize}

Assessment issues can be divided into those concerning {\em user-centered
assessment}, and those concerning {\em technology assessment}.

User-centered assessment involves the testing the operational feasibility
and functional adequacy of system, with regard to its intended
user population, and to gauge user acceptance and the socio-economic impact
of the system in a real-life situation.  More specifically, it concerns
such questions as:

\begin{itemize}

\item How is market research performed and how are user requirements defined?

\item How are different classes of user taken into account?

\item What are the factors in the general environment the system is
      used in that significantly affect its usability?

\item How is evaluation of adequacy in terms of user satisfaction to be 
      performed?

\item What are the methods of quality management?

\item How are usability, scalability, portability and maintainability ensured?

\item What are the migration paths and methodologies for porting applications?

\end{itemize}

Technology assessment refers to activities undertaken to measure the
performance of the technology(ies) used within a system, with an
aim to: assessing progress development within a project; establishing
the suitability of a given technology to a set of stated goals; to
compare methods and results with a view to stimulating technology
transfer; to establish the potential of a given technology, e.g. in
terms of portability, scalability and integratability with other
technologies.  More specifically, it concerns such questions as:

\begin{itemize}

\item How is progress measured?

\item What are the technological baselines and starting conditions required
      for progress measurement?

\item What are the criteria for choosing a given technology?

\item How is the fulfilment of functional specifications determined?

\item How does the technology relate to the intended application and
      user population?

\end{itemize}

Given the emphasis of FP-4 on user-led rather than technology-driven
projects, it is important that technology assessment should be tied in
with user-centered assessment.  One needs to be able to assess, and
predict, what kinds and what properties of technology lend themselves
best to different applications and user groups.  Little is to be gained
from measuring technology performance if this either has no bearing on,
or even an inverse relation to, the user adequacy of systems employing
the technology.  Taking an example from another area, one could measure
the technological performance of (formula 1 racing) cars in terms of how
fast they go; but this would not necessarily be the primary or most
appropriate technology measure for domestic motorists.  One of the
primary aims of technology assessment is to provide the means to
abstract away from individual users, and to gauge the applicability of
systems or system components to other types of user.

\subsection{Provenance of the Interim Report}

This report is based around, and includes material from, four trigger
papers written on various aspects of evaluation and assessment, and responses
to the papers.  The trigger papers, and one reaction paper, are

\begin{itemize}

\item Sparck Jones \& Crouch: General Technology Assessment

\item Netter: Technology Assessment for Written NL Applications

\item Steeneken: Speech Technology Assessment

\item Adriaens: User-Centered Assessment for RTD in Language Engineering

\item King: Reactions to G. Adriaens

\end{itemize}
Further reactions to the trigger papers were received from Yorick Wilks,
Louis Pols, Steve Young, Steve Pulman, Karen Sparck Jones and Klaus Netter.
The report also contains the results of a questionnaire composed
and circulated by Rob Gaizauskas, to which there were some 10
responses.

The editors have made free use of this material, and have recycled some
of it more or less verbatim.  However, the overall way in which this
material has been used (or misused) remains the responsibility of the editors.

The report divides into four other sections. The first introduces
terminology to be used in the report. The second discusses user-centered
assessment, followed by an extensive discussion of technology
assessment. In section 5., some recommendations are given about the kind
of organisation and infrastructure required for carrying out
user-centered and technology assessment, both at a project internal
level and for the purposes of comparative evaluation.

\section{Background and Terminology}
\label{termin}

In this section some of the terminology used throughout this report 
is introduced.  This is, unfortunately, quite important.  When there
is little agreement on common terminology, as is the case in evaluation,
the definitions invoked are usually a good predictor of a person's
final views.  From the terminology here, one can predict a strong 
predisposition towards a form of comparative technological assessment that 
takes user-centered considerations into account.  For wider ranging
discussions of terminology and illustrative examples, see 
Galliers \& Sparck Jones 1993, and EAGLES 1994.

\subsubsection*{Systems}

A {\em language engineering application system} or just {\em LE system} is a
set of software {\em components} 
constructed to permit a user
to carry out some language-related {\em task} or {\em
function} in a specific real-world
environment. We will assume that all pilot applications funded under the
FP-4 call in Language Engineering will be LE systems.

\subsubsection*{Tasks}

Just as an LE system can typically be broken down into a number of
individual components and sub-components, so can the task (or function)
it performs be broken down into a number of tasks and sub-tasks.  
Generally, we can talk of a system or system component as {\em implementing}
a task.  However, it is important to realise that this need not always
be the case, and that task structure need not always exactly match
system structure.

For example, a (hypothetical) connectionist language analyser might not
have any separately identifiable components implementing the tasks
of morphological analysis, syntactic parsing and semantic interpretation;
these are all wrapped up in one monolithic system component.  From
one point of view, say that of someone designing a modular language
analyser, the system still performs these separate tasks.
But one could well imagine the connectionist system designer arguing that 
these do not correspond to any genuine tasks, as witnessed by the lack of any
separate system components.  For the purposes of evaluation, we
want to avoid the ensuing factional arguments.  By separating tasks
from components, we have the option of super-imposing different
task structures on top of a single system structure in order to promote
a variety of meaningful comparisons and evaluations.

Thus, identifying task structure rather than system architecture is the
first step towards defining an evaluation framework.  System architecture
will of course normally provide much useful information about the
appropriate task structures to employ.

A task is a {\em mapping} from an {\em input} object or set of input objects
to an {\em output} object or set of output objects.  Tasks can be largely 
defined by these input-output mappings. Some examples of language engineering 
(LE) tasks are: translation (texts in, texts out); parsing (sentences or word
lattices in, parse trees, or labelled bracketings or dependency structures 
out); speech recognition (sounds in, lists of words or word lattices out).

\paragraph{Visibility and Transparency:}
It is helpful to distinguish {\em user-visible} and {\em user-transparent}
tasks.  User-visible (or user-significant) tasks are those where
both the input and output objects have some kind of direct functional
significance to a system user.  User-transparent tasks are ones where
the input and/or output are of no direct interest to the user, so that
typically a user-transparent task is part of a wider user-visible task.
In a machine translation system, the translation task will be
user-visible: the input and output texts are both significant objects
from the user's point of view.  But parsing, carried on as part of translation,
is likely to be user-transparent: the user does not know and
does not care about parse trees.  Whether a task is user-visible or 
transparent can depend on the user as much as on the task.  In a multi-lingual
information access system, translation may well be user-transparent:
the user does not know and does not care about what language the information
originated in, and thus only cares about the output of translation.

\paragraph{Individuating Tasks:}
As the parsing task illustrates, input-output mappings do not completely
define tasks. This is particularly a problem for intrinsically
user-transparent tasks, like parsing, since the inputs and/or outputs
are theoretical objects.
Depending on one's perspective , one can hold 
different views on what kinds of theoretical objects should be
involved. This is perhaps even more marked in the case of semantic
interpretation, where the output might variously be discourse representation
structures, situation schemata, quasi logical forms, montagovian logical
forms, etc.  So strict identity of input or output objects is not
required for task identity.  What does seem to be required is that
\begin{enumerate}
\item There is some degree of shared information between input objects
and between output objects.  There should be a relatively trivial
purely syntactic transformation that converts one input (or output) 
representation to another input (or output) representation, perhaps with
considerable loss of information.  For example, one can convert word lattices 
to word lists (by just taking one path through the lattice), and word
lists are a trivial form of lattice.  Similar transformations, again with
often considerable loss of information, are possible for different syntactic
and semantic representations.
\item The input and output objects have basically the same functional
role within similar larger task structures.  Thus, whatever the details of 
parsing output, it is typically has the role of input to a further
task of semantic interpretation, which in turn might feed into some
sort of inference task.  Eventually, these chains of roles have to bottom out
in obviously user-visible objects (sentences, texts, answers, etc.).
\end{enumerate}
Fortunately, there is fairly widespread agreement about what constitute
the main language engineering tasks, despite theoretical variation.

\paragraph{Task Attributes:}
Tasks can have different {\em attributes}, some of which are
\begin{description}
\item{\em I/O Attributes:}  
There is often room for variation in the nature
of the inputs and outputs to tasks.\footnote{How much variation a task
can survive before becoming a completely different task is touched on above.}
For LE tasks, {\em linguistic features} that can vary (for different tasks) 
include:
\begin{itemize}
\item The language  of the input or output (e.g. French, English);
\item The subject area (e.g. weather reports, financial
      newswire stories);
\item Text type (e.g. newswire stories, technical manuals, spoken dialogue,
      dictation);
\item Text length (e.g. sentence, paragraph, article);
\item Spontaneity (e.g. read speech, spontaneous speech);
\item Channel conditions (e.g. telephone, wide band);
\item Accent (e.g. native, non-native, regional);
\item Speaker (e.g. speaker dependent, speaker independent).
\end{itemize}
In addition to this, {\em theoretical features} can have an impact on
the task, e.g. what syntactic or semantic framework is being employed.

\item{\em Internal Objects:}  
Tasks can rely on internal objects (or perhaps
concealed input objects), such as grammars, lexicons, language bigrams,
statistical preference weightings, etc.  Often, these internal objects
can be altered to take account of differences in I/O attributes. For
example, in a parser different grammars and lexicons for different languages
or even for different theoretical frameworks.

\item{\em Depth, Accuracy and Robustness:}  
A task can vary in the degree of depth,
accuracy and robustness with which it maps inputs onto outputs. A task
is done shallowly if certain details of the output representation are
not captured, and is done deeply if they are.  A task is done
accurately to the extent that the details of the output object that are
represented are or are not ones that are really there.  A task is done
robustly if it can produce some sort of output for any input, rather than
just failing on some inputs.  

One might wish e.g. to consider different settings of the depth, accuracy and 
robustness attributes for parsers to be employed in different settings.
Accuracy and robustness are often also attributes of components 
or systems implementing tasks.  But along with depth they are definitely
also task attributes: one can deliberately aim to implement a shallow,
robust parser, regardless of implementation.

\item{\em Efficiency:} 
The speed, and the resources consumed in performing
a task are often important attributes relating to the user adequacy
of systems.  Efficiency is often more of a component (i.e. implementation)
attribute. But some tasks are also inherently time or resource consuming,
regardless of implementation, while others are inherently efficient.

\item{\em Task Decomposition:}
There may be more than one way of decomposing a task into constituent
sub-tasks.  Different task decompositions may be more or less
appropriate to different constraints on the linguistic features, the depth
and accuracy, and the environment within which a task is to be performed.

\end{description}

Of course an LE system may perform many tasks that are not linguistic
--- for instance it may provide a user with an interface to select
files to be processed for translation, to store the output of translation
operations in files, and so on. Proper performance of these tasks may be
as essential to the success of the system as carrying out its linguistic tasks.

\paragraph{Applications:}
A useful piece of terminology is that an {\em application} is a task
plus a domain, where the domain covers such things as subject matter and text
type (more generally, the universe of discourse). Thus, we might have 
two applications for the same task, e.g.
a translation system for weather reports and a translation system for
airline enquiries.  Similarly, we can have two applications for a
similar domains, e.g. translation of airline enquiries and automatic
answering of such queries.

\subsubsection*{Environments}

Tasks, and the systems that perform them, do not exist in isolation.
They are embedded within a wider environment.  For user-transparent
sub-tasks within a system, the environment is determined by the
sub-tasks surrounding it.  For user-visible tasks, the environment is
determined both by the other user-visible and -transparent system
tasks surrounding it, but also by the properties of the users, e.g.
degree of training, degree of urgency in getting the task done,
importance of getting the task done exactly right, the other non-system
based tasks they need to perform, and so on.

Environments thus display a concentric, onion-like organisation, where
what constitutes a system, task or component at one level may correspond
to part of an environment at a lower level.  One can consider an entire
{\em setup} (computational system plus end-users) within the wider environment
or corporate or national economies.  One can consider a computational
system within the environment constituted by its end-users.  One can
look at a major task components in a system (e.g. translation, document
retrieval, speech recognition) in the context of other major task components
in the system.  One can view lower level tasks (parsing, coreference
resolution) within the environment of the broader linguistic tasks
they contribute to (e.g. translation, summarisation, message routing).
And so on.

It is important to realise that it is not just systems that have environments.
Any task or component (including non-linguistic ones)  within a system also 
has an environment. 
Evaluative principles stressing the need to match a task or its implementation
to its environment do not apply just at system level.  The nature
of the environment might change as one descends to ever deeper levels
within the system, but the same general principles apply.
It is therefore a mistake to draw a completely hard line between
user-centered / system  assessment on the one hand, and technology /
individual component assessment on the other.

\paragraph{Environment Attributes:}
To match a task to its environment, we need to compare task attributes
and {\em environment attributes}.  By and large the categorisation of
task attributes carries over to environment attributes.  Thus the
environment can determine the kinds of inputs and outputs expected
of tasks (which language, subject domain, text type, etc); the depth,
accuracy and robustness required of the task; the efficiency required.
In other words, environmental attributes impose requirements on the
task, and one needs to check whether the task (and/or its implementation)
has the appropriate attributes to meet these requirements.

It is useful to identify environment and task attributes
that can vary, and those that (for a given evaluation or setup) are fixed.  
The variable attributes will be called {\em environment variables} and 
{\em task (or system/component) parameters}.  For example, suppose the
intended users of a translation system may variously want quick and
dirty translations, or clean but slow translations.  Then the
environmental attributes of depth, accuracy and efficiency constitute
environmental variables.  If the system is to cope with these varying
requirements, the system/task attributes of depth, accuracy and efficiency 
should also be variable, and constitute system/task parameters.

In many cases, it may not be so obvious which task parameters need
to be varied to cope with different values of environmental variables.
A {\em grid} evaluation methodology naturally emerges. 
This involves carrying out different
{\em runs} of a system or component, varying environmental attributes (e.g.
language, subject domain, text type) against different settings of
task parameters.  Depending on what the evaluation is meant to show,
one might wish to vary more than just the environmental variables,
e.g. if one wishes to investigate how a system of component would fare
outside its intended environment.

Environment attributes tend to percolate down as one increases the
granularity of tasks being considered.  Thus a system environment
attribute concerning speed of processing will typically surface as
a similar environmental attribute for components or sub-tasks within
the system.  This means that user-centered aspects will usually need
to be taken into account when considering the environments in which
system sub-components or sub-tasks are to be evaluated.  So even
user-transparent sub-tasks are subject to user-centered considerations.

At this point the task/component distinction is useful.  There may be
two reasons why environmental attributes are not matched by
system attributes. 
First, the components implementing the system task have not been
coded with sufficient generality or flexibility; in this case
the system may be doing the right thing, but not well enough.  Second,
the task itself does not have the right kind of attributes; in this
case, the system may be trying to do the wrong thing. (See validation and
verification).

\subsubsection*{Users}

Users, who have been somewhat neglected in most speech and language assessment,
come in two major varieties
\begin{enumerate}
\item
The people who actually use the system on a day-to-day basis 
(end-users)\footnote{There may be different types of end-user, e.g. teachers
and students using some kind of educational system.}, and
\item
The people who are paying for the system (procuring users)
\end{enumerate}
The two, and their respective interests, may or may not coincide.  End-users 
will probably want something that makes their lives easier, more
interesting or more enjoyable.  Procuring users will be more concerned with
productivity increases, and may not much care whether the system improves
the lot of the hands-on users, so long as it makes them more
productive (though one would hope that former leads to the latter).

Users are the prime determinants of system environment variables.
It is therefore necessary to obtain user profile data, and in this
regard we should distinguish further categories of user
\begin{itemize}
\item present users, accessible for collection of user data
\item prospective users, from whom such data may or may not be obtainable
\item idealised users, to who future users will hopefully correspond
\item individual users: if a system is being tailored to a specific individual
their idiosyncratic needs should be taken into account
\item collective users (e.g. home, professional, or mobile): most users
within a collective group will exhibit broadly similar profiles, though
with exceptions.
\end{itemize}

\subsubsection*{User Tasks}

User Tasks or User Roles can be seen as an abstraction from users as
individuals or groups for the purpose of modelling certain tasks or
roles that a prototypical user has to perform in a given scenario or
domain. These tasks can be defined on the basis of a work-flow analysis
and can form the basis for measuring improvement of effectiveness in the
performance of the tasks.

\subsubsection*{Validation and Verification}

Assessment comprises verification and validation.
The difference between verification and validation is succinctly summarised
as
\begin{itemize}
\item Verification: are we building the system right?
\item Validation: are we building the right system?
\end{itemize}
Verification is thus a form of assessment that gauges how far a system
or system component fulfills its functional specification, i.e. the
degree to which it carries out its specified task. Validation is a
form of assessment that checks that a system actually meets genuine
needs and requirements.

Validation ultimately comes down to {\em user validation}; namely that
a system, the tasks it aims to perform, and the components implementing
those tasks all contribute to meeting the needs of its intended users.
To carry out the validation, it is vital to have a profile of the
users and to determine the environment variables and range of possible
settings these profiles give rise to.  Having determined these, one
way of subjecting a system to user validation is by carrying out
a grid evaluation, varying environment variables alongside system 
parameters.  But this kind of laboratory test is no substitute for
real user-acceptance testing.

Thanks to the concentric onion-like organisation of environments,
one can abstract away, to some extent,  from the needs of specific users and
talk about task or technology validation.  Ultimately, a task or
technology only has validity if it can produce results that are beneficial 
to some user.
So one cannot have `pure' task validation, independent of any
user requirements. But some tasks and technologies are so far from 
being user-visible that it is impossible to provide direct user validation.  
Instead, one needs to validate them against the environments set up
by other tasks, to ensure that the task in question
really does deliver the kind of results that are wanted for other tasks.
User requirements still have an attenuated influence at this
level of validation, through the effects they have on surrounding tasks.

\subsubsection*{Technology}
A technology is a task for which there exists one or more alternative
components implementing it, and where substantial variation of internal
task parameters to match possible environmental variables is permitted.  
For the
task of parsing, there exist a number of implementations (chart parsers,
shift reduce parsers, statistical parsers).  These can be parameterised
by grammars, lexicons, statistical preference ratings, etc to tailor them
to the different environmental variables of language, subject domain, 
text-type, etc.  Thus we talk of parsing technologies.

Technologies are re-usable in different situations.  They can be deployed
to implement a task within a wide variety of environments.  A technology
has user validity if it can be deployed within some system meeting
genuine user needs.  The more such systems it can be deployed in,
the greater the re-usability and validity of the technology.

\subsubsection*{Technology Assessment}

There is a tendency to draw a strong contrast between `technology' and
user-centered assessment.  What is usually meant by technology assessment
here is the evaluation of mainly user-transparent tasks or components
of a system, in a way that either abstracts away from (or more likely
ignores) user-centered factors. Not all sub-tasks or components within a
system meet the criteria above for being a technology; some can be
quite specific to a particular application or system. Therefore,
the term technology assessment can be somewhat misleading: (user-transparent)
task assessment might be better terminology.  However, the term is entrenched,
and we will continue to use it to mean a more general form of task assessment.

In the light of the preceding discussion of onion-like
environments, it would a mistake to view technology assessment as
being entirely divorcible from user-centered factors.  A task or technology
has to be assessed in the light of environmental factors, and these
will include factors ultimately stemming from system users.  

To be sure, many previous efforts at technology assessment have proposed
abstract measures (such as matching labelled bracketings, predicate
argument structure, word error rates etc) and applied them to system components
without regard to user issues.  This is not necessarily incorrect, merely
incomplete. For the assessment to enjoy proper validity, one also
needs to match the results of these measures up against identified
environment attributes for the task.

In other words, there is a pressing need for many of the technology
assessment measures currently being advocated to be subjected to
(environmental) and user-centered validation.  One needs to establish
that the measure being used do correlate with useful properties,
especially from the user's point of view.  Technology validation is
just as important as technology verification.

\subsubsection*{Internal Assessment}

Internal assessment refers to user and technology verification and validation
that is carried out solely with regard to the particular needs of an 
individual project.  Internal assessment typically has a diagnostic as
well as an evaluative element.  One does not only wish to establish
the degree to which the system does the right thing in the right way.
One also wants to identify, during the project lifecycle, where things
are going wrong, what needs to be done to correct them, and whether attempts
at correction have been successful. 

Internal assessment will normally make use of evaluation data specific to the
needs of the project: either the user needs, or the technological needs.
User-specific data will be determined by the nature of the intended
application: the subject matter (weather reports, airline queries,
aerospace technical documentation), and the task (translation, information
access, document management).  Given that applications will vary from
project to project, this user-specific data may not be directly
amenable for use as comparative evaluation data

Technological evaluation data will depend on the precise instances of 
technology used within the project.  One might for example have data 
for assessing
parser performance, which involves sentences paired with their intended
parser output.  The details of the parser and its output might be quite
specific to the project.  So again, the technological data may not
be directly amenable for use as comparative evaluation data.

There are some areas where the border between internal and comparative
assessment becomes a little grey.  Important properties of any system
are maintainability, adaptability and portability, which all pertain
to the ability to use the system outside of its initially intended
specification and environment.  Changes in environment might result from: 
different user groups, different domains, addition of further task 
requirements, and so on.  Assessment of these properties
gauges how the present system compares to slight variants of
itself, and is a form of comparative assessment.

\subsubsection*{Comparative Assessment}

Comparative assessment involves taking different systems and comparing
either their system-wide performance, and/or the performance of
individual components / task competencies.  Systems and components can vary 
according to their task, domain, user profiles/environments, the 
technologies they employ, and their implementation and task/system 
decomposition.  More generally
\begin{description}
\item{\em Environment:}  This has several sub-variables, applicable in 
different cases:
\begin{itemize}
\item User profile (system assessment)
\item Overall surrounding task (component/sub-task assessment)
\item Domain, text type, etc
\item Channel conditions, speaker accent, etc
\item Other non-linguistic factors
\end{itemize}

\item{\em Task:} One can vary the task or system goals

\item{\em Task Attributes:} These include
\begin{itemize}
\item Depth, accuracy, robustness and efficiency
\item Grammars, lexicons, language models etc
\item Task decomposition
\end{itemize}

\item{\em Implementation:} Which covers such things as hardware platforms, 
operating systems, programming language, code, etc
\end{description}
One can fix, vary or ignore combinations of these factors 
to get different  kinds of evaluation, not all of which are
strictly comparative.  Varying things at the 
system level, one can have, e.g. 
\begin{itemize}
\item ARPA: 
The ARPA MUC and TREC evaluations are characterised by having the system 
environment and task fixed, while varying task attributes and implementation.
\item Flexibility: 
For an individual system one can vary environment factors such as 
user profile, domain, etc to see how flexible a system is.  Assessing
domain independence may be important here.
\item  Usefulness:
By fixing just the user profile,  one can assess how useful
various systems are for a particular user group
\item Portability: 
e.g. by varying the hardware platform, operating system.
\item FP-4(?): Let everything vary
\end{itemize}

An important question is whether it is possible to have a meaningful
comparative evaluation when all factors are allowed to vary, since the
systems to be developed under FP-4 certainly will vary along all dimensions.
A second question is what benefits such a comparative evaluation will bring
in the near and/or long term to system users.  These questions are addressed
at greater length in section~\ref{TAC}.  Briefly, the benefits of
comparative evaluation are
\begin{itemize}
\item Cross-Fertilization:\\
By comparing different systems, technologies and implementations, projects
can learn from and exploit best practices adopted in other projects.
\item Maintainability and portability:\\
Comparative assessment inevitably extends a system's performance beyond
its originally intended domain, fostering the properties of
maintainability and portability.
\item Identifying promising technologies.
\end{itemize}

\subsubsection*{Evaluation Data}

The core evaluation activity involves providing a system or system 
component with some input and comparing its output with the results
expected.  The comparison may takes several forms besides a simple yes or
no depending on whether the results were as required.  One could measure
the time and resources consumed in obtaining the results; if the
actual results differ from the desired ones, one could attempt to quantify
the difference in terms of both depth and accuracy; one could even
measure the amount of effort involved in porting a system from one
subject domain to another.  But for all these modes of comparison,
evaluation data is required.

Evaluation data comes in three forms
\begin{enumerate}
\item Test, or input data\footnote{
In the ARPA evaluations, `test data' referred to the combination of what we
have here called input and output data.}
\item Answer, or output data
\item Training data
\end{enumerate}
Training data may not always be required, but for any system employing 
statistical  methods it is likely to be essential.  Very often, old
input and output data can be used as training data.

Requirements on evaluation data are that it be {\em realistic}, 
and {\em representative} (Sparck Jones 1994).  
For test data to be realistic,
it must be the kind of input data that the system or component would
actually receive in real use.  For it to be representative, it should 
contain instances from the full range of input data that would be received.
Moreover, the data should either
be easy to acquire, or if not then widely reusable for other purposes.
This issue especially affects answer data, since acquiring it may
often involve laborious hand (or semi-automatic) annotations of texts
or other linguistic objects.

A distinction should also be drawn between diagnostic data and adequacy
data.  Diagnostic data may be set up very carefully to pinpoint particular
points at which a system is failing.  Adequacy data is used to check the 
degree to which a system or component performs as required, without 
necessarily attempting to pinpoint the sources of any failings.
The requirements of realism and  representativeness are somewhat 
weaker for diagnostic data than for adequacy data.

In collecting evaluation data for assessment, a decision must be made
about the level of:
\begin{enumerate}
\item  {\em granularity} at which systems will be evaluated,
e.g. at the level of
    user-significant tasks only, or at some level of tasks
    that are user-transparent)
\item  {\em generality} at which systems will be evaluated (e.g. how much do 
    we vary the
    linguistic features of the input/output data we provide/expect relative
    to the features of the data in the intended application; e.g do
    we evaluate the system against data from different languages, different
    domains etc.).
\end{enumerate}
These two dimensions are orthogonal: the decision to evaluate at a high level
of granularity is independent of the decision about which linguistic features
of the input/output data are to be generalised, if any, and to what extent.
The choices made will in part depend on whether internal or comparative
assessment is being contemplated.

\subsubsection*{---ilities}

There are a number of `ilities' constituting desirable characteristics
of computational systems (ISO 9126)
\begin{itemize}
\item Functionality: should satisfy stated or implied needs

Covers: suitability, accuracy, interoperability, compliance, security

\item Reliability: should behave in a predictable and consistent way

Covers: maturity, fault tolerance, recoverability

\item Usability: minimise effort required to use system

Covers: Understandability, learnability, operability

\item Efficiency: cost effective use of resources

Covers: time behaviour, resource behaviour

\item Maintainability: ease of making modification

Covers: analysability, changeability, stability, testability

\item Portability: ease of transferring from one environment to another

Covers: adaptability, installability, conformance, replaceability
\end{itemize}
The main purpose of evaluation and assessment is to (a) foster these
desirable properties within individual projects, and (b) to lay the basis
for greater satisfaction of these properties in future projects.

\subsubsection*{Measures and Metrics}
Assessment metrics need to be specified at three levels
\begin{enumerate}
\item (Qualitative) Criteria:\\ what kind of thing is to be measured, e.g.
translation quality, productivity improvements, \ldots
\item (Quantitative) Measures:\\ how the criteria are to be measured, e.g.
amount of post-editing required, increased throughput of documents, \ldots
\item Application Methodology:\\ how the measures are to be applied in
a uniform and consistent way.
\end{enumerate}

\section{User-Centered Assessment}
\label{geert}

This section\footnote{Most of this section is taken, more or less
verbatim, from Geert Adriaens' trigger paper.} addresses user-centered 
assessment.  As such it is
primarily, though not exclusively, concerned with project internal
evaluation; the next section on
technology assessment takes up the issue of comparative assessment.

\subsection{The Project Level}

\subsubsection{Evolutionary Life-Cycles}

The classical project development life-cycle consists of the following 
major phases:
\begin{enumerate}
\item requirements analysis and definition (in LE terms: preparation stage)
\item system and software design  (in LE terms: development / implementation
verification  and testing  stage)
\item delivery and maintenance   (in LE terms: demonstration, documentation,
                   exploitation stage)
\end{enumerate}
In its simple form, this model has advantages, but also major drawbacks. One
advantage is that it allows for imposing clear milestones (assessment points) 
at the  transition from one major phase to another. 
A major drawback is that users, if consulted at all, are brought in either too 
 presupposes that the different stages are independent, can nicely be finished 
before proceeding to a next stage (with an all-or-nothing delivery at the 
end), and hardly need feedback-plus-change iterations

To overcome these drawbacks, a controlled evolutionary development model with
incremental deliveries seems to be more appropriate. Briefly, the approach 
consists of breaking up an intended system into smaller functional units, 
and applying the prepare/develop/verify/demonstrate stages to each of these
units. These units constitute intermediate deliveries that evolve into the 
final system after several iterations. The approach has the following 
advantages:
\begin{itemize}
\item
Maximal user involvement: for each relevant system increment, users can and 
should participate in  all stages (from analysis to test), and trigger a 
number 
of iterations before accepting a component. Users are no longer forced to 
state all their requirements (which they may only know when they see a
 concrete 
system component!) beforehand, and to accept a fully developed system of which 
they can at most suggest some cosmetic changes at the end.
 
\item
Complete analysis, design, build, test and document in each step; much tighter
interwovenness of all phases (for instance, one can no longer afford to 
postpone
writing a test plan corresponding to requirements, or to postpone 
documentation 
writing forever)

\item
Maximal adjustment flexibility: if mistakes are made, they can be detected 
and corrected early; if certain design or tool options change (given the 
rapid evolution in the software world), they can be incorporated more easily
 
\item
Inherent need for open-endedness and extendibility: all increments must be 
fitted seamlessly, and hence maximally modular, compatible and interoperable 
(which should also improve reusability)
 
\item
Bottom-up early-result orientation, not top-down last-minute software 
development orientation
\end{itemize}
 
The model also entails certain risks that require careful consideration:
\begin{itemize}
\item
The number of iterations at different evolutionary stages must be kept under 
control (say, maximally three). For instance, users may tend to keep changing 
their minds as they see different proposals for interface solutions. At some 
point, they should formally accept a solution so as to allow the project to 
continue its course, and not go around in circles.
 
\item
Feedback loops require a solid management of change (by experienced project 
managers), and a carefully designed verification and validation approach; 
regression tests, for instance, become crucial (see below)
 
\item
The evolutionary approach implies a lot of activity at the micro-level of a 
project, but the risk is that one is unable to see the wood for the trees. 
Hence, it must be controlled; one way of doing this is by still imposing the 
classical development milestones (end-of-analysis, end-of-design, 
end-of-laboratory-development, end-of-alpha-test, end-of-beta-test) on the 
evolutionary cycle, although the point in time may be kept flexible. 
And of course, as for any type of project, resource constraints 
(time, money, people, tools) largely determine what is possible for each step.
 
\item
The evolutionary model also has implications for an external evaluation and 
review process; rather than only having major mid-term and end-of-term 
reviews, 
less extensive but more frequent checkpoints or spy-points may have to be 
imposed on the project. Although at first sight this looks like a
lot of overhead, it allows for timely adjustment and better overall project
results.
\end{itemize}
 
We will take this controlled evolutionary development model as the starting 
point
for discussion of user centered assessment.  The model is especially designed
to promote the ease and timeliness of user assessment.

\subsubsection{User Profiles}
User can either be end-users or procurers.  They may be characterised
individually, collectively, prospectively or hypothetically.  But in all cases,
information profiling users is a necessary starting point for user-centered
assessment.

A user profile is a concise description of a user's abilities, interests, 
preferences, etc. as confirmed by collectible data or measurements. 
One can look at it on the basis of different types of data:
\begin{itemize} 
\item
Factual, objective data about users\\
These determine the {\em implied needs} of the users.
\begin{itemize}
\item
Background factors (education, training, experience, knowledge, 
language abilities, skills, etc.)

\item
Physiological factors (vision, hearing, touching, dexterity, speed)

\item 
Psychological/cognitive factors (learning, memory, attitudes, beliefs, 
expectations, mental models; the latter constitute the framework of 
concepts, objects, actions, structures, tasks, metaphors etc. the user 
has in mind for subsequent visualisation and use in interaction with a 
computer)
\item
Application interest and use factors (frequency of use, goal in using)
\end{itemize}

\item
User opinions, preferences, subjective data (obtained by interviews or 
questionnaires)
 
These data correspond to the {\em expressed wants} of users and need not be 
consistent with the objective data. Confronting both may already help in 
finding the "real" needs; the next type of data can also be very helpful 
in determining these real needs.
 
\item
Measured or observed data, obtained by recording how the user actually 
performs 
in interaction with the computer for a particular application

These data correspond to {\em demonstrated needs} and wants. Practically, they 
can be obtained by video taping, or better still by internal recording 
mechanisms added to an application. This can range from simple modules 
gathering usage statistics   to sophisticated software tools that record all 
on-screen user manipulations in a kind of movie. The latter could be 
complemented by video taping to also record off-screen activity. Of course, 
not 
every application should go so far, but the module  gathering usage 
statistics  
should not represent a major effort and still be a very useful instrument in 
trying to objectively determine what the users really want on the basis of 
their 
behaviour with a system (in order to produce improved subsequent versions)
\end{itemize}

\subsubsection{System Attributes}

User profiles determine system-wide environment attributes and variables.
The system attributes that should be matched against environment
attributes should be identified, and success criteria and measures for
meeting the system attributes must be set up.

Two very broad system attributes are usability and integratability.  These
attributes can be refined, and tested in the following manner.
 
\paragraph{Usability} is related to the following issues:
\begin{itemize}
\item
Level of personal ability to enter training courses (if any) for the product.
\item
Training time required to attain a pre-determined level of productivity with 
the product 
\item
The specified amount of work to be produced by a person so trained
\item
The rate of errors made by a trained user, operating at the normal work rate
\item
The opinion of the users as to how well they liked the product
\end{itemize} 
Certain aspects can clearly be measured: training
 time, level of productivity (what percentage of tasks was executed in how 
much 
time?), number of errors made. For each of these, criteria can be set up 
(average, worst case, best case), using, for instance, data from experienced 
users and multiplying these by a certain factor (say, execution time should 
stay 
within a factor two times average experienced user execution time).
 Note that the profile of the tested users is an important background element, 
and also that independently of the task results, a likability questionnaire 
should be handed to users (because they could perform extremely well on a bad 
system -- which they actually hate --given external pressure such as fear of 
losing one's job).
 
Once the criteria are set out, very concrete tests can be set up that address 
the different usability issues, along the following hypothetical lines:
\begin{quote}
\ldots after 75 minutes of training, 40 typical users should be able to 
accomplish 80\% of the benchmark tasks in 35 minutes with fewer than 12 
errors\ldots
\end{quote} 
 Such a test could very well figure on the list of acceptance tests of the 
buyer/procurer of the system. Acceptance is part of the validation process, 
which will be looked at in the next section.

\paragraph{Integratability} is an important attribute that pervades all 
stages of the development cycle of a system or system increment. At the 
lowest level (or the earliest stages) in development it refers to 
compatibility and interoperability of subcomponents:
do they exchange and mutually use information properly? do they support each 
other's functions? At the highest level, it refers to the potential of a 
system to be integrated seamlessly in the end user environment.  
Since users are central in this discussion, this is the kind of integration 
that needs further refinement, the more so as NLP products are not noted for 
their smooth integratability. 
Unfortunately, little experiential data exists about large-scale integration 
efforts and field tests, probably partly because many NLP solutions tend 
towards the custom-built solution rather than the off-the-shelf product, 
or simply never make it as a product. There are also no standard procedures 
for checking integratability at the higher levels, i.e. integration in a 
complex user environment. As network access issues (client/server 
architectures, information highway access) become more important, there is 
also an added factor of complexity: standalone solutions are history; there 
is always an integration aspect. The following are some rough ideas on
 how integratability could be controlled.
\begin{itemize}
\item
Next to user profiles on the microlevel, corporate organisation profiles 
(companies, institutions, etc.) or subdivision profiles (documentation 
division, translation division, EDP division) could be drawn up at the 
macro level to get an idea of the global  environment in which an 
application will be integrated
\item
Information and work flow analysis and the effect of new technologies on 
these in terms of changes in processes, structures, tasks; changes in user 
profile requirements (new types of occupation, need to retrain or hire 
specialised personnel); changes in computer facilities, etc. could be 
investigated
\end{itemize}

\subsubsection{Testing and Acceptance}

The following test types occur during development of a system or system 
increment; they are ordered chronologically, by decreasing involvement 
of system creators and increasing involvement of users, and by increasing 
importance of integration matters in ever-broadening contexts:
\begin{enumerate}
\item 
component test (unit, module)
\item 
integration test (in the laboratory)
\item 
alpha test (system or system increment test by other people than the  
developers, typically prospective end users, possibly still in a controlled  
environment)
\item   
acceptance test (a formal test initiated by the prospective procurer/buyer of 
a 
system or system increment, typically at the installation site; for 
custom-built 
solutions, this is a final test in the test chain for a system or system 
increment)
\item  
beta test (in case of a development which has to become a product at large --- 
as opposed to a custom-built solution -- further field testing of a completed 
system with selected prospective users) 
\end{enumerate}
At each level of testing, different iterations following feedback and test 
results may be needed, i.e. regression tests must be carried out as a way of 
monitoring progress. Regression testing presupposes a rigid approach in which 
test data are well-defined, and evaluation tools exist to determine changes 
from one test run to another. Planning of regression tests is also notoriously 
difficult, because it is unpredictable whether and how fast a system will 
converge to a stable state. Care should be taken to foresee time for these
regression loops inside the different testing phases.
 
In the context of EC projects with partners in different countries and ever 
more complex cluster structures, it may be important to apply acceptance 
tests internally to the project. After all, from an industrial perspective, 
the typical project structure is one of a prime contractor (with final 
responsibility), and a number of partners required to deliver subcomponents. 
Hence, the prime contractor should have a formal acceptance procedure for 
outsourced components as part of his quality assurance plan. This may not be 
a trivial task, because different companies may have different quality 
assurance plans and acceptance criteria; moreover, the prime contractor 
cannot impose his own approach on the other partners. To avoid conflicts in
this respect, it may be a good thing that partners make explicit their 
quality assurance approaches; if a partner does not have an in-house 
approach, one could decide beforehand to adopt the approach of another 
partner (typically the prime contractor). In these situations, adherence 
to overall standards (such as ISO-9000) would be a solution, but we are 
not that far yet. In the meantime, however, ANSI/IEEE standards do exist 
for different stages of a development cycle; they provide a backbone or
checklist of activities and procedures that should not be overlooked.
One such standard is ANSI/IEEE 1012-1986 on Software Verification and 
Validation Plans.

Acceptance testing requires a test plan, for which the following issues
are relevant:
\begin{itemize}
\item
What are the organisations involved and their responsibilities in the 
acceptance test?
\item
How can requirements be traced to test cases?
\item
How will the acceptance process be administered?
\item
Are the test cases and test procedures complete?
\item
How will error reporting and error analysis be done?
\item
What will be the location, testing approach, facilities, equipment, training 
for the tests?
\item
What are the time and money resource implications of the tests?
\end{itemize}
Next, test design is important (and should in fact be considered at 
requirements and specifications time);
\begin{itemize}
\item 
What are the items to be tested for?
\item
What are the objectives and constraints for each subtest?
\item
How can the test design and test cases be traced to system requirements 
(i.e. are they already considered in the requirements and specifications 
documents?)
\item
What are the supporting tools required for each test?
\item
What are the expected inputs and outputs of each test case?
\item
What are the initialisation and stopping conditions for testing?
\item
What are the concrete criteria to say that the system passed the test?
\end{itemize} 
Finally, considerations are needed for concrete test procedures and for 
documenting all test aspects:
\begin{itemize} 
\item
What test procedure corresponds to the test design and test cases?
\item
Who (what particular user(s)) will do the testing where?
\item
What are the required pretest conditions in the test environment?
\item
How are test results reported?
\item
In case of problems, how are problems reported and what is the procedure 
for handling them?
\end{itemize} 
 To help users in respecting a certain formality in the approach, standard 
test approach forms and problem reporting forms should be used. They will also 
greatly facilitate the correct reporting and documenting of the test 
activities.
 
An issue that is often overlooked in test planning and execution are
difficulties arising because the outcome is totally unpredictable. One cannot 
foresee what problems may crop up, and what the turnaround time will be for 
fixing them. This is also related to the difficulty of regression testing: one 
can never be sure that a problem or its solution has no side-effects on other 
aspects of the system. If a serious problem occurs, further testing may become 
impossible, and deadlines may be become impossible to keep. To cope with this,
sufficient time must be reserved for testing; also, testers should signal 
problems right after finding them, so that the procedure for solving 
them can be set in motion immediately.
 
\subsubsection{User Interfaces}

The importance of user-interfaces, and the difficulty of assessing them,
should not be overlooked in user-centered evaluation.  However good the
underlying system, a poor user interface can make it practically unusable.
Adriaens's trigger paper contains many valuable comments on interface
design.  From the point of view of testing, a major difficulty arises from
the fact that it is difficult to precisely reduplicate test circumstances.
Within an interactive system, the user's knowledge and circumstances
inevitably change over time, so that one cannot usefully repeat individual
tests.

\subsection{The Project Cluster Level}

Since different sets of users are involved at the project
cluster level, it is not immediately obvious what role user-centered
assessment has there, or even whether it is possible.

But most of the reasons for pursuing comparative evaluation do persist
at the user level: e.g.  cross-fertilization through the recognition and
adoption of promising approaches and technologies; promotion of
maintainability and portability by evaluating a system outside of its
immediately intended environment to gauge the ease with which it
could move between environments.

One possible mode of user-centered comparative evaluation would be to try the
user group from one project out on a system from another project
having a broadly similar,  though not necessarily identical, functionality.
As well as providing further information about learnability, it is possible
that users will find features in other systems that they would like
to see incorporated in their own.

Another mode of evaluation would be to exchange subject domains between
projects in the cluster (assuming that suitable evaluation data
can be made available), to assess the flexibility and portability of
systems.  However, if systems are especially unportable, the cost
of changing domain may be prohibitive. Domain change could either go in tandem 
with, or be separate from, exchanges of user groups.

Finally, and again probably assuming that systems can change domains,
integratability could be addressed by actually trying to combine
either parts of systems or entire systems.

All of these modes of comparative evaluation have an over-optimistic
ring to them.  The problem with user-centered comparative evaluation
comes down to the fact that user groups may simply be too heterogeneous
to permit much meaningful comparison.  This being so, there is a danger
that user groups within projects will be resistant to
this form of comparative evaluation.

What appears to be feasible at the present stage, however, is to draw on
the expertise and experience that different projects have to and will
build up during project internal validation and to encourage a strictly
bottom up development of standards. Although user-centered evaluation
will always be geared towards the individual users, it is quite likely
that users will be interested in profiting from experiences with
validation methods gathered in comparable projects and applications.

For this purpose, the user-centered evaluations methods applied by the
different projects should be made publically available and/or collected,
compared and processed by some independent institution. What could be
distilled from these data are abstractions from comparable evaluation
scenarios, evaluation metrics etc. In the medium and long run this could
result in a kind of evaluation library containing user models, abstract
scenarios, evaluation methods, test suites, tools, metrics etc., which
have been successfully applied in user validation.

Clearly, this will not result in a comparative evaluation proper,
however, such a collection and its continuous improvement and refinement
could lay the foundation for more broadly accepted standards in the area
of user-centered evaluation.

\section{Technology Assessment}
\label{TAC}

\subsection{Technology and Comparison}

Purely user-centered comparison, viewing systems as black boxes, is a
doubtful proposition.  User groups (and systems) tend to be too heterogeneous 
to
make meaningful comparisons feasible.  However, if one looks at the structure
of different systems, one usually finds that they have tasks and components
in common.  Of course the same task/component in two different systems
might have quite different attributes in terms
of user-visibility, depth, accuracy, robustness, efficiency, language,
text-type, etc etc.  Any comparison between the task performance in the
two systems will need to take these differences into account, and also
try to relate performance to user attributes.

In other words, assessing underlying technology (i.e. task and
sub-task performance) is a natural way of pursuing comparative evaluation.
However, just because some of the tasks being compared are user-transparent
does not mean that user considerations are absent from technology assessment.
As emphasised in section~\ref{termin}, user-based environment factors percolate
down to influence user-transparent tasks.  Technology assessment permits
a natural abstraction away from specific user requirements, but abstracting
is not the same thing as ignoring.

It is vital that one does not abstract away completely from
user-centered considerations.  `Pure' technology evaluation metrics
require user-centered validation.  There is little benefit to be gained
if the technology metrics bear no relation to the ability of a system
to perform its chosen task for its chosen user group.  Technology
must be evaluated relative to the environment in which the technology
is to be used, and this brings in a reference to user requirements,
albeit indirect.

User-centered validation of technology evaluation metrics has received
scant attention.  The metrics proposed in the course
of such things as ParsEval and SemEval (labelled bracketings,
predicate-argument structures, co-reference relations, etc) make
sense from a technologist's point of view.  But beyond intuition
(which may well turn out to be reliable), not much is known about how
these metrics correlate with the performance of systems, taken from a user's
point of view.  Although highly unlikely, it is conceivable that
that there is no correlation.  Rather more likely is
that for some tasks there is an inverse correlation.  To take an admittedly
fanciful example, success in identifying co-reference relations may have
an inverse correlation with the adequacy of spell-checkers; this
might reflect the fact that co-reference imposes few if any constraints on
spelling, and that effort which would have been more profitably directed
elsewhere has been wastefully devoted to co-reference processing.

If user-centered validation is to be taken seriously, this means that
the initial stages of any comparative technology evaluation exercise
would need to be devoted to it.  The aim would be to propose a
number of candidate technology metrics, and to measure them on a variety
of systems in parallel with carrying out direct user-centered evaluation.
Validation would consist of correlating the two evaluations, in the
hope of finding which kinds of technology are best suited to which
kinds of task.  It is possible that this would require a number of
iterations in order to find the most informative technology measures,
and this could well be a long-term undertaking.

After initial phases of user-centered validation, benefits
would start to flow.  One being the prospect  of being
able to make informed choices about the choice of technologies in
the initial design of a system for a given task. On the assumption that
technology assessment can be carried out using standardised test data,
users and systems designers can identify appropriate technological tools
without first having to perform user-specific evaluation, where the
collection of test data is likely to be expensive.  This is not to say
that user-specific testing can be dispensed with; far from it.  Merely
that in the initial feasibility and design stages it may be dispensable.
Another benefit is that users and systems designers would have
a much better idea of the ways in which a system can be modified
and adapted to deal with different tasks.

Comparative assessment is of potential value to user groups because
it promotes:\footnote{
These point are intended to address some doubts raised by Maghi King, who
has said ``there seems to me to be a tension between
the FP-4's emphasis on involving specific user communities
throughout the life-time of a project and and any attempt to impose some sort
of comparative technology evaluation. This is because user communities have
their own very specific interests, and I think it will be hard, if not
impossible, to  convince them to collaborate in any evaluation initiative
which they perceive as falling outside their direct  interests. Thus,
for example, I suspect that they will not be interested in using common
test data unless that data matches their own concerns.''}
\begin{itemize}
\item Cross-fertilization\\
One of the bye-products  of the ARPA MUC and TREC comparative evaluations
has been cross-fertilization between different systems.  If one system
employs a technique that is shown perform particularly well on a certain
sub-task, then the other systems in the evaluation have tended to make
use of that technique to improve their own performance.

Admittedly, the ARPA evaluations all compare systems performing the same
task, on the same subject domain, for the same hypothetical user group.
This makes cross-fertilization particularly easy.  But systems designed
for a more disparate range of tasks are still likely to have at least
some sub-tasks in common (e.g. spell-checkers and information retrieval
systems will probably involve word segmentation and morphological processing).
Comparative evaluation means that funding user groups can be more confident
that the most appropriate components have been employed to perform various
sub-tasks.

More generally, the competition that comparative evaluation engenders means
that technologists developing systems are kept on their toes and
don't become complacent.  Given that most users will probably not have the
technical expertise to detect such complacency, this is bound to be in
their interests.

\item Maintainability and portability\\
Over the course of time, the precise tasks that a system is called upon
to perform tend to shift.  It is important that the system
be maintainable and portable in order to deal with these shifts in use.
One way of promoting this kind of flexibility is to test and evaluate
a system by consideration of tasks and application domains lying outside
its immediately intended areas.  A properly set up comparative evaluation
can do this, since components of a given system will be run on
data from different domains, and where the data arise in response to
different sets of user requirements.

\item Identifying promising technologies\\
Of longer term interest is the fact that comparative evaluations should
serve to identify promising technologies.  This may not be of immediate
interest if a given system has been set up in such a way that it is
committed to a particular, less than optimal, technology.  But for
developing future systems, the information may be valuable.
\end{itemize}

Additionally
\begin{itemize}
\item
Comparative assessment involves a degree of abstraction away from
specific tasks, domains and user groups.  Some kind of general
technology assessment is required to achieve this abstraction.
\item
Performing meaningful technology assessment entails the identification
of relevant environment and user attributes.
\item
Any technology measures used must have a user-centered rationale.
\item
User-centered validation has not previously been carried out to any
satisfactory degree.  Therefore, an important initial component of
any comparative evaluation exercise for the FP-4 should be
this kind of validation.  This involves proposing general technology
metrics, applying them to individual systems, and comparing the results
with user-centered assessment of the systems.
\item
One consequence of this is that comparative evaluation is unlikely to
lead to a league table saying which of the participating systems is
best.  Rather, it should lead to a grid identifying what it is
about different systems that makes them well suited to their task.
\end{itemize}

\subsection{Constraints on a Comparison Exercise}

Substantial comparison exercises have been undertaken by ARPA.
While having certain flaws, these exercises have been of benefit.
However, it is unclear whether comparative evaluation under FP-4 can
or should follow the same path.  The main feature of the ARPA exercises
is that, while growing from small beginnings, they have been primarily
top-down.  Systems have been constrained to operate on the same application
(and for the same hypothetical user group).  But within
FP-4 there will be a wide variety of applications and user groups.
A top-down, ARPA style approach is not applicable to all the FP-4
projects in their entirety.

One possibility would be to set up a small ARPA-like evaluation project
under FP-4.  However, this would (a) exclude remaining FP-4 projects
from comparative evaluation, and (b) does not sit well with emphasis
on users in FP-4 --- the evaluation task would doubtless involve
a large degree of artificiality. Nevertheless, amongst many technologists
there is enthusiasm for a top-down approach.

Another way of tackling things would be to build an evaluation exercise
bottom-up from existing FP-4 projects, hopefully building on the work
done for internal project evaluation.  Not all of the FP-4 projects
need be involved.  Indeed, it would be more practical to start with a
limited number aimed at connected, though not identical, applications.
In what follows, we assume that a primarily bottom-up scenario is more in tune
with the aims of FP-4.

If a more bottom-up approach to comparative evaluation is to succeed, it is
important that participation be relatively painless and inexpensive.
Especially so, given that in its initial stages the benefits of
participating in a comparative evaluation may be slow to accrue.  This in
turn means that the evaluation exercise may start off rather small but
grow, so that the exercise should have a structure that allows for
development over time.

In reality, the distinction between a top-down and bottom-up evaluation
exercise is somewhat artificial.  A `left-corner' framework, combining both
top-down and bottom-up features is preferable.  In this, one tries to
build up from individual projects, but in order to do this a certain
amount of standardisation (in terms of the types of evaluation data used,
and the measures employed) needs to be imposed top-down.

More specifically, constraints on the exercise are that:
\begin{enumerate}
\item
The exercise should be incremental for the future, with respect to data,
tests and
experience: it should be possible to roll the materials, experience and results
of earlier stages forward into later ones.
\item
The exercise should have low entry and working costs to allow and encourage
participation.
\item
The evaluation should address tasks with multilingual and multimodal
(spoken and written language) aspects, falling within the areas covered
by approved LE projects.
\item
The materials should be cheap, including both working data e.g. corpora and,
more importantly, the evaluation data with `answers' for chosen
tasks.
\item
The materials should as far as possible be reusable, or multipurpose, for
other S\&LP
research.
\item
The evaluation program for computing performance measures should be relatively
easy to provide and apply.
\item
The evaluation structure should allow both technological
and user-centered evaluation.
\item
As far as possible the comparative evaluation exercise should sit on
top of, and make use of, project internal evaluation.
\end{enumerate}
In addition, as argued previously the evaluation exercise should, in
its initial stages if not throughout, address itself to the question
of validating technology metrics in user-centered terms.

\subsection{Braided Evaluation}

A {\em braided} evaluation
structure is a candidate meeting the requirements set out above.  The
braid model starts from the observation that tasks of
any substantial complexity can be decomposed into a number of linked sub-tasks.

Some of these tasks will be user-significant, others will be
user-transparent\footnote{The latter were called {\em jobs} by Sparck Jones
and Crouch in their trigger paper.}
For example, a document authoring and management system
may perform a number of distinct user-significant tasks (spelling correction,
grammar correction, dictionary/terminology access, document retrieval),
and these may depend in turn on a number of user-transparent tasks
(word segmentation, morphological analysis, parsing, selection of index
terms for document retrieval, etc).

Sub-tasks within an overall task will be linked by language objects,\footnote{
Not all tasks performed by a system will be linguistic, e.g. low level
file access, graphical interfaces, etc.  We are implicitly confining
attention to speech and language related tasks here.}
which might be texts, sentences, parse trees, predicate-argument
structures, phoneme lattices, etc.  A given sub-task will have one or
more input objects and one or more output objects.  The input and output
objects provide obvious material for black-box evaluation of the sub-task
(which will become glass-box evaluation if the sub-task is further
decomposed).

Sometimes, all sub-tasks within an overall task will be linked in a simple,
linear, nose-to-tail structure.  But more often, the sub-tasks will
exhibit a branching, or even braided structure.  The more complex
structure reflects the fact that in a complex task (like creating a
document, say), there are different routes through the task structure,
and different options that one might employ or ignore on different
occasions.

Different overall tasks will often have overlapping sub-tasks.  The more
similar the overall tasks, the greater the degree of overlap.  When
two or more tasks are decomposed and put together, what emerges is
a braid-like structure.  At some points, the task decompositions will
diverge, maybe to partially converge at a later point, or maybe never
to join up again.  A given sub-task may have alternative decompositions,
so that although it has the same input and output objects, it may pass
through quite different intermediate objects under different decompositions.

The point of decomposing tasks and putting them together in a braid structure
is that it allows one to identify natural common evaluation points.
These are where two or more overall tasks have common sub-tasks, even if
the sub-tasks themselves are not necessarily composed in the same way.
However, care must be exercised here.  The fact that two tasks share a common
sub-task does not mean that those tasks have identical input and output
properties; merely comparable ones.  It is quite possible that two tasks
sharing a common sub-task pursue the sub-tasks to different levels of
detail.  For example, a parsing sub-task may be required to give only
a shallow analysis in the context of one overall task, while a deeper more
detailed parse tree is required for the other.

To recapitulate:
\begin{itemize}
\item
The braid consists of sequences of (linguistic) tasks, linked through common
(language) objects --- e.g. sentences, texts.
\item
There are natural evaluation points, not only at the end points of the whole
braid, but at other intermediate points referring to one or more of the
preceding member tasks.
\item
These evaluation points allow for a variety of different kinds
of evaluation
\begin{itemize}
\item
When a sub-task is user significant, then user-centered evaluation metrics
can be applied.
\item
When a sub-task is specific to a particular kind of task, e.g. retrieving
documents matching a query or filling in slots in a standardised template,
then metrics specific to that particular kind of task can be applied
(e.g. precision, recall, etc).
\item
When a sub-task can be seen as an instance of a general S\&LP task
(parsing, morphology, word recognition), then general technology
metrics can be applied
\end{itemize}
\item
A braided structure provides a flexible framework, allowing comparison
of different tasks where possible.
\end{itemize}
It should be emphasised that the braid model is intended as an architecture
to support
evaluation, and not as a system architecture.  While sub-tasks may
well correspond to separate processing modules in a system architecture,
this certainly does not have to be the case.  The tasks define
competences to be evaluated and not system components; it is possible
for a task to be distributed across several components, and also for
several tasks to make use of the same system component.

\subsubsection{Common Resources for Braided Evaluation}

A braided evaluation structure allows for comparative
and individual evaluation of different systems at different levels
(user-centered, task-specific, general technology).  It has the
flexibility to incorporate new tasks and systems as it goes along.
Unfortunately, it does not make comparative evaluation a simple
and trivial matter, requiring no central resources or effort.
While every attempt should be made to bring project internal and
comparative evaluation as close together as possible, it would be
unrealistic to expect a comparative evaluation to run itself,
with the only impetus coming from within individual projects.

One of the important insights of previous attempts at evaluation on a
large scale has been that it always turned out to be a very cost and time
consuming enterprise. Both cost and time can be reduced if methods, test
data, and interpretation results can be shared by a larger community.
For many institutions, the possibility of sharing in such an infrastructure
for an evaluation scheme has, by itself, proved to be sufficient attraction to
voluntarily undergo an evaluation.

\subsubsection*{Obstacles to Shared Infrastructure}

However, the sharing of such an infrastructure is hampered by various
factors, such as diversity of application tasks, domains, system
architectures, text types and---crucially in a multi-lingual
environment---diversity of languages.  A braid model is not, in itself,
a solution to these problems.

\paragraph{(a) Tasks}
Research politics in the past (and possibly
present) has favoured a diversification of technologies and
applications, in the sense that projects had better chances for funding
when they suggested not only a different solution and technology for a
known problem, but also when they suggested a new and interesting
problem. ``No parallel research'' has lead to a vast
amount of diverse technological developments in NLP, which on the
surface are hardly comparable and are difficult to assess against each
other, mainly because they are claimed to serve their purpose only
within specific and different applications.

Task specific assessment of technologies will thus be feasible
only if there is a large enough number of systems performing the same
tasks. Task oriented assessment of technologies would require that
the contributions of a module to an application could be factored out in
a reliable way.

\paragraph{(b) Domains}
Even if more than one project works on the same
task, it is almost invariably the case that they will not involve
the same subject domain. Typical examples here are small to medium scale MT
systems, which are often constrained to specific domains.

What is more frequently found though is different tasks being
for the same domain. However, here the obvious problem arises,
that domain specific test data typically do not carry over between
different tasks if they are annotated in a task-specific
way.

\paragraph{(c) Evaluation Data}
Evaluation data has to be chosen that is
sufficiently representative for the task or application to be assessed.
Normally, this problem is circumvented by collecting a sample which is
large enough to fulfill this requirement.  In very few cases
{\em test suites}, with fully controlled and more compact test material,
have been constructed. To our knowledge, none of these test suites have been
applied to evaluation as opposed to diagnosis; test suites are generally
designed to meet criteria other than user-centered evaluation.

Large corpora by themselves, however natural and representative,
 do not suffice as the basis of an automatic
assessment procedure, since they only provide the input but not the
output against which the performance has to be assessed. What is
required in most cases is therefore some kind of {\em annotation} to the
corpus, providing answer data.  Depending on the type of technology
assessed, these annotations can range from tagging, phrase structure
annotations, etc., to content classifications in some abstract way.

The problem is that most reference corpora and annotations of this kind
are tailored towards the assessment of some specific technology or
task. Again, to some degree this appears to be justified if the
text sort or the domain are very specific to the application. However,
this also means of course that the efforts for collecting and annotating
corpora have to be duplicated, even for those cases where annotations
could be shared at least in part.

\paragraph{(d) Languages}
In a multi-lingual R\&D environment
comparability across languages represents another important factor.
Trivially, parallel multi-lingual corpora could play a role in the assessment 
of
MT systems, although adequacy of MT systems is hardly ever tested by
comparing system output with a predefined translation on the level of
corpora.

Parallel test corpora in different languages with comparable annotations
could become relevant, if portability of systems from one language to
another is to be assessed.

\subsubsection*{Layered Corpora}

What appears to be required to address the issues and problems
mentioned above
is the provision of an infrastructure which allows an open competitive
assessment on the basis of predefined reference material which supports
glass box evaluation of tasks and technologies wherever this is feasible.

This involves not only the development of an appropriate organisational
framework, but also some considerable effort in constructing corpora
serving as the basis of the assessment. Klaus Netter's trigger paper
argued that these should be assigned
layered annotations at different levels of abstraction matching the
envisaged intermediate reference levels for glass box evaluation.
These would correspond to the input and output language objects linking
tasks in a braided evaluation structure

These annotations could include all kinds of information, such as
morpho-syntactic tagging, word sense disambiguation, phrase structures,
relational structures, semantic representations including resolved
references, but also specific annotations for a range of envisaged
tasks. What these reference levels are would have to be agreed on
by the groups involved. Ideally these corpora should also be constructed
in parallel for different languages.

The idea of corpora with layered annotations (rather than
different annotations distributed over a wide number of corpora) would be that
projects can enter a competition on the basis of a corpus without
being constrained by either their application or by the specific
architecture of their system. Even if it would not necessarily provide
the assessment basis for some specific application, the technology of
some subcomponents or modules could still be evaluated. Of course,
systems would also not be forced to follow a strict architectural setup,
since they could equally well be assessed at only those subsets of the
reference levels, which are relevant for the respective applications and
which are employed in the individual architectures.

While it might be possible to provide a single layered corpus, this would
probably be a mistake.  First, one would have to exercise great care in
selecting a corpus that lends itself equally well to a wide variety of
different tasks.  Second, a single corpus would not provide the means
necessary for assessing the domain independence of systems.  One would
therefore hope to be able to build up a handful of layered corpora,
uniformly annotated.

\subparagraph{Other Resources}
Provision of layered corpora is liable to be expensive, and some ways
of reducing the cost by building on project internal corpora are suggested
below.  But in any case, more than corpora will be required to support
comparative evaluation using common data.  This includes further linguistic
data and resources, as well as non-linguistic resources

Extra linguistic resources needed might include: lexicons,
terminological databases, or  statistical domain models.
Non linguistic resources might include: the data held in exemplar databases,
logical axiomatisations of world knowledge, and so forth.
Not all the extra resources would be required for all systems being
evaluated.  However, some kind of database or information access system
would require, in addition to a corpus of possible information requests,
some kind of database containing the answers to the requests.  A document
authoring aid would probably not require this kind of resource, but if
the system was intended to allow the incorporation of material from other
documents, it would require data about those documents.

Software tools for scoring results, and possibly also the semi-automatic
annotation of answer data would also be required.

\subsubsection*{Exploiting Project Internal Resources}

Any project that tackles internal evaluation seriously is likely to build
up evaluation data in the form of layered corpora, lexicons, terminological
databases, databases, etc, as described above.  If at all possible, it
makes sense to build on this kind of test data to provide material for
comparative assessment.

However, even if individual projects can be persuaded to part with
their own, hard won test data (in exchange for equally hard won
data from other projects, perhaps), this alone is not enough.
The problem is not that the data will be specific to a particular domain.
In fact, this is an advantage, since evaluation across a range of
domains is a good way of ensuring that systems do not contain ad hoc
domain-specific short cuts.  The problem is that the test data will only
be appropriate for certain tasks and sub-tasks.  It will typically not be of
much help in evaluating those parts of other systems dealing with
quite different tasks applied to the same domain.

For example, a mono-lingual information access system for airline
reservations simply will not provide the kind of test data required
for evaluating the multi-lingual aspects of a multi-lingual access
system for airline reservations.  Nor is it likely to provide the
kind of annotated test data required for evaluating some authoring tool
dealing with an airline reservation domain.

Test data provided by individual projects is thus a starting point for
building up shared evaluation resources, but it has to be built upon.
It needs to be extended to cover tasks not relevant to the system from
which the data originated.  In addition, the project data itself needs
to be subjected to some kind of quality control: a test suite of half
a dozen questions and their expected answers does not even provide
a good starting point for building shared resources.

To further facilitate the re-use of project internal test data, a degree
of top-down imposition is desirable.  If a core of technology metrics
can be decided upon in advance (e.g. labelled bracketings,
predicate-argument structure, co-reference relations, word sense
identification), then projects can be encouraged to produce internal
test data appropriate to these kinds of metric.  This does not mean
that projects cannot also develop test data annotated towards further
requirements of their own, nor that they are prevented from arguing for
the inclusion of these further annotations as part of the common test data.
One needs to retain flexibility and room for development in what should
constitute common test data.

\subsubsection{Relation to ARPA}

Standards for technology assessment of written NL applications are only
gradually emerging. The most prominent instances of practical
applications in this area in recent years have been the MUC and TREC
conferences for message understanding and text retrieval respectively,
and possibly schemes such as ParsEval and its upcoming successor SemEval
(some of which will be part of MUC-6) for the testing of broad coverage text
analysis components.

For MUC and TREC the main cornerstones of the assessment methods were
black box evaluation, the usage of naturally occurring corpora as the
test material, and the employment of ``recall'' and ``precision'' as
evaluation metrics. Both conferences led to advances in the field not
only by challenging competitions but also by exchanges of experiences of
technologies. An important factor on the success side appears to have
been that these conferences, although sponsored by ARPA, were not
limited to ARPA sponsored projects, but organized on the basis of an
open competition.

The main criticism that has been levelled against the evaluation methods
in these conferences is that it did not support the development of
strategically relevant promising technologies (this is somewhat
truer of MUC than TREC). Since what was evaluated
by these conferences was task performance rather than technologies in
the narrower sense, they supported task and application oriented
short-cuts to some degree.

In contrast to the task oriented MUC and TREC, Parseval was an approach
which attempts to evaluate at the module level by using a benchmark
method. Its basis was corpora which were annotated by some syntactic
structure agreed on by a representative number of experts. The output of
analysis components was measured against these annotations.

The main criticisms of Parseval are on the one hand that the syntactic
annotations had to be very abstract to allow mapping of the output of
different analysis systems onto these annotations. Systems which
over-performed and produced more specific output where thus at a
disadvantage relative to systems which would produce just about the
required result. Second, phrase structure analyses still favour a certain 
theory
specific convergence on what is assumed to be an accepted structure.
SemEval is similar to Parseval in these respects, e.g.
over-performing systems with a deeper semantic
understanding will still be at a disadvantage, and even if predicate
argument structure is more abstract it will still not be completely
theory neutral.

What appear to be the main limitations of current evaluation methods, if
viewed from the point of technology assessment are the following
aspects.

Many of the performance evaluation methods, such as employed in MUC and
TREC, are task specific. Thus, it is practically impossible to
compare technologies unless they are integrated into an environment
which tests the respective task.

There is little methodology for evaluating the portability of
technologies. This partially results from the concentration on
application specific assessment, but partly also from the limitation to
pure task-independent assessment. In a multi-lingual environment the
latter criticism can be applied in a similar way to the porting of
technologies from one language to another. Important factors for the
strategic advancement of technologies are thus practically ignored at
present.

\subsection{An Illustrative Evaluation Scenario}
\label{ta-scen}

This section expands on a scenario for a braided evaluation exercise
outlined in an earlier trigger paper.  We should
emphasise that this is for illustrative purposes.
We are not proposing that this particular scenario should form the
basis for an evaluation exercise set up under the FP-4.
It would be possible to use the scenario in this way; but it is preferable
to pursue a more bottom-up approach to deciding on an evaluation
structure.  In order to involve as many LE projects as possible
in evaluation, it is better to wait and see what tasks and projects are
actually to be worked upon, and then build the task structure up from that.
In this scenario, we merely try to show how various types of evaluation
can be combined, with the possibility of validating them against each other.

The task scenario centres on document retrieval and translation,
assuming that document requests couched in one language may need to
be searched against document files in other languages, and that retrieved
document titles in one language may need to be translated to others.
This is something that could very easily form a sub-task within
various LE projects: e.g. those dealing with document management and
authoring and those dealing with information access.  In addition, the
translation aspects will have points in common with machine translation
projects.  One could therefore envisage a considerable range of further
tasks and sub-tasks being added to the braid.  But here we deliberately
consider only a relatively simple task structure.

We will start by summarising the main path through the task structure:
\begin{verbatim}
   L-OBJ 0             Request sentences for documents

   TASK 1              Select and translate index terms for
                       document retrieval

   L-OBJ 1         >>  Search term lists

   TASK 2              Document retrieval

   L-OBJ 2        >>   Retrieved documents

   TASK 3              Translate retrieved document
                       titles and/or abstracts

   L-OBJ 3        >>   Translated document titles/abstracts
\end{verbatim}
This gives a very coarse breakdown of the task structure, and all of the
tasks could be sub-divided into further sub-tasks.

Task 1 is likely
to be user-transparent: in many cases, users will not be interested in
index terms selected for file searching by the system, only in the documents
retrieved as a result of them.
The pair L-OBJ 0 and L-OBJ 1 therefore will not provide suitable material
for user-centered evaluation.  However, it will provide material for
a task specific evaluation, which e.g. checks on the soundness and completeness
of the index terms selected (soundness: proportion of correct indexes selected,
and incorrect indexes not selected; completeness: proportion of total number
of  correct  indexes actually selected).

This task-specific evaluation needs to be treated with care.  What counts
as the correct index terms may very well depend on the nature of the
retrieval system carrying out Task 2.  So direct, task-specific comparisons
between systems employing different retrieval systems may very well
not be possible.  This illustrates one of the problems with task-specific
evaluation: the nature of a task may depend on its surrounding context,
so that even two apparently identical tasks may not be directly comparable
when their contexts differ.

Depending on the nature of the systems being tested, Task 2 may not in fact be
regarded as an S\&LP task, although it may very well include
some sort of language processing.  The pair L-OBJ 1 and L-OBJ 2 provides
material for evaluation of the retrieval system.  But from the point
of view of the task structure as a whole, the pair L-OBJ 0 and L-OBJ 2
is perhaps of more interest.  The transition between the two is very likely
to constitute a user-visible task, so that user-centered assessment
is possible.  This would involve not only comparing the language
objects, but e.g. assessing the extent to which the document retrieval
improve the speed and quality of whatever wider task the combination
of Tasks 1 and 2 is embedded in.  The language objects also permit a relatively
context neutral task-specific assessment of overall retrieval, since the
form of the desired results are liable to be largely independent of constraints
imposed by the following translation task.

Task 3 may very well be entered from other directions, without first
going through document retrieval.  One could envisage user, task and general
technology measures being applied to Task 3.  In addition, the combination
of Tasks 1--3 is similar to that of Tasks 1 and 2, permitting task-specific
and user centered assessment.

The coarse task decomposition above provides little space for technology
assessment.  We will illustrate how it can be included with a
further decomposition of Task 1.  There are various ways in which
one might map requests onto sets of index terms, ranging from keyword
spotting to full syntactic, semantic and pragmatic processing.   This in
part depends on whether requests are short, one sentence questions, or
paragraph long statements of interest. Hence,
the decomposition of Task 1 braids into a number of different paths.
There is one path going through e.g.: (1) word segmentation and morphology;
(2) parsing; (3) semantic interpretation; (4) reference resolution and
pragmatic processing; (5) identifying key concept combinations;
(6) producing index terms.  Some of these may be decomposed further; thus
(2), (3) and (4) may very well involve further disambiguation tasks. Various
steps on the path may be missed out.  One might go straight from word
segmentation and morphology to either the production of index terms
or identifying key concept combinations, or do the same from parsing,
semantic interpretation, or reference resolution.  Or one may sidestep
the entire path by doing simple keyword spotting.  Since most of the
tasks on the path (1)--(5) correspond to standard linguistic functionalities,
one can apply general technology assessment measures, e.g. those
identified by Parseval and SemEval.  However, different systems
may perform these tasks to different depths of analysis and levels of
detail. It is therefore important that the metrics are able to give some
measure of analysis depth, as well as of correctness at the chosen
level of detail.

Validating technology assessment will involve taking (i) the presence or
absence of a particular task, (ii) the depth of analysis to which the task
is performed, and (iii) the proportion of correct to incorrect results,
at the chosen level of detail, and correlating these with the
various dimensions along which Task 1 can be deemed to succeed at
a task-specific level, and Tasks 1 plus 2 at the task and user level.

Possible amplifications of this chain, assumed so far based entirely on
written language, are an alternative entry to Job 1, with spoken requests;
and an alternative exit after Job 2, when retrieved document texts are
processed to extract key concepts: depending on how this extraction task
was defined, it could be a natural extension of the evaluation after the first
stage.
Note, however, that own-language document retrieval should probably
 not be treated as a separately evaluable task for short requests.
This is because, on current test findings, it does not require extensive
S\&LP for typical retrieval situations and would thus not be an attractive
subject for S\&LP evaluation: the presumption is that request translation,
especially for some languages e.g. compounding ones, could be linguistically
more exacting and not be limited to simple dictionary lookup.
Own-language retrieval would be done only to provide evaluation data.
However the retrieval task with translated requests could be varied
with a `baseline' version designed to study translation effectiveness in
its own right, and a `souped-up' version where actual retrieval performance
exploiting all search resources like weighting could be assessed.

In relation to speech processing evaluation in particular, various
speech input conditions are possible for Task 1, and there are also
speech input and output possibilities at Task 3.  At this sort of level,
one can address question of how deep (full parsing etc) and shallow
(keyword spotting) approaches to index selection fare with different
word accuracy rates.  Does keyword spotting work better with poorer
word accuracy in recognition since it is inherently less brittle, or does the
extra context imposed by deeper linguistic processing allow one to recover
more easily from word errors?

The other subpaths allow for less or more exigent S\&LP, e.g. according to
whether title/abstract translation is undertaken.
It is possible to enter and exit the chain at different points, but never
without a rational job or task and evaluation step for this.

\subsection{Spoken Language Issues}

Since the illustrative evaluation scenario above is somewhat biassed 
towards text processing, this section\footnote{This section is derived
from a contribution of Herman Steeneken's.} raises some  issues more
specific to the evaluation of spoken language systems.

Integration of speech products in
daily applications is growing (e.g. dictation systems, banking and
travel information inquiries, and dialogue systems as used for training
and education).
Three topics of particular interest in the evaluation of the present
state-of-the-art speech technology:
user appreciation of spoken language systems; assessment of technology
modules;  and the interaction between spoken language and natural language
systems. 

 Many development
centres and industries are interested in comparing their in-house systems
with current state-of-the-art systems.  To facilitate this, 
data bases used for assessment and the description of the procedures,
scoring metrics and results need to be disseminated within the community.  
For most of the assessment procedures it is crucial that test date are
``unseen'', so that systems are not adapted or tuned by making use of
the test data. This requires a continuous stream of new test data.
This may be feasible for certain conditions but for a wide range of
applications availability of these type of data will be limited.
Nevertheless for representative tests, data for various languages,
speech types and recording and transmission conditions are required. A
central coordination point might be required as well as general accepted
classification criteria.

A point of consideration concerns the items to be assessed and possible
assessment methods. Items to be assessed include:

\paragraph{1. Applications}
\begin{description}
\item[a1 Command and control:] generally related to the operation of ``hands
and eyes busy'' systems. Speaker dependent recognition and a small text
output vocabulary are used. The major issue is total system performance
and error correction.

\item[a2 Document creation:] requires large vocabulary recognition and robust
error correction.  Systems may be adaptive to speaker, environment,
vocabulary, speaking rate and other variables. This adaptation requires
specific assessment methods.

\item[a3 Information retrieval:] may be a public service (telephone speech,
speaker independent, dialogue structured, speech understanding). Total
system performance (does the user get the required information
efficiently), user appreciation and the performance of different
technologies are main issues for assessment.

\item[a4 Tools for disabled:] includes a wide variety of specific applications.
Speaker dependent recognition for retarded speakers which may deliver
deteriorated speech are the major issues.

\item[a5 Training and education:] can be characterized by command and control
type applications but also (such as with second language teaching)
combined with written text applications.

\item[a6 Security control:] includes speaker identification, speaker
verification, and language identification. The assessment is very much
application related and generally requires specific data bases.

\item[a7 Translation systems:] include multilingual speech I/O (in general 
large
vocabulary applications). Many aspects are to be assessed such as
input/output modules and the translation module itself.

\item[a8 For simulation:] (including virtual environment applications) all
aspects are covered by a1, a2, a3, and a5.
\end{description}

\paragraph{2. Modules}
\begin{description}
\item[m1 Speech input:] can be divided into isolated/connected
word recognition (for which many methods are developed) and large
vocabulary recognizer assessment. For large vocabulary recognition in
general ``untouched'' test data are required and specification of the
vocabulary and lexicon type.

\item[m2 Speech output:] for pre-recorded speech may rely on existing speech
intelligibility tests. For text-to-speech the dimensions intonation and
speaker and style variation should be added. This normally requires
subjective tests.

\item[m3 Transmission:]  in adverse input/output conditions
(back-ground noise, telephone speech) transmission of the speech signal
requires a robust physical
specification of the conditions.  Standardization, by making use of
simulation, offers a possibility.  Multilingual use requires a variety
of identical (language specific) data bases. To make this feasible a
``universal'' application can be used (i.e.  travel information) as well
as extension of existing robust corpora to other languages.
\end{description}

\paragraph{3. Standardization} 
\begin{description}
\item[s1 Bench Marks:] in order to compare the performance of systems 
evaluated within
different projects (a thread that pulls together projects within a
robust programme such as sponsored by the CEU) should include
standardized measuring methods, sharing of tests material, and uniform
scoring methods. Also bench marks or reference conditions can be
considered. Robust bench marking allows new developers or industrial
competitors to scale the performance of their system within the range of
performance of state-of-the-art systems.

\item[s2 Data bases:] should be (within commercial confidence) accessible by
other projects within the same frame work. General coordination should
include merging of tools between related projects.
\end{description}

Possible assessment methods include:
\begin{description}
\item[User appreciation:]
Human factor studies normally require subjective measures based on
queries opinion scores, rating, (pair-wise) comparison, etc. These
techniques are well developed as well as the statistical tests which are
applied to analyze the responses of subjects. However, the application
of these techniques to quantify the user appreciation of NLP products is
not well developed.  

Therefore, specification and development of
controlled repeatable user appreciation tests is required.

\item[Total system performance]:
\begin{itemize}
\item benefit vs. human performance
\item     successful trials
\item     handling time
\item     error analysis
\item     ease of use
\end{itemize}
Total system performance quantifies the technical success of a
system (does it work well). It can be expressed in measures like
percentage of successful trials, handling time, etc. Some of these
measures are a spin-off of a robust user appreciation test in which some
of these objective measures are included.  Additionally the benefit of a
system versus human performance (do we really need such a system) are
items to include. Human performance also offers a bench mark for system
evaluation.  Especially in combination with adverse conditions human
performance tends to be superior.

\item[Technology assessment:]
Progress during development, competitive comparison and progress
estimation are normally quantified within a specific application or
technology (e.g., large vocabulary recognition). For technology
assessment the conditions are carefully controlled (laboratory
conditions) in order to obtain reproducible results. Depending on the
purpose of the assessment a black versus a glass box approach can be
selected.  Some test designs are competitive which pushes the technology
foreword. The test paradigm should include the share of results and
methodology after the test (used with ARPA and SQALE) in order to keep
participants interested.  Interaction between a human and a system may
fail. Therefore, the method of error correction and an analysis of the
errors occurring in man-machine dialogues is an aspect to be addressed.

Realistic adverse conditions can be included in laboratory tests
(artificial, reproducible, inexpensive) or in field tests
(representative, uncontrolled, expensive).  In general development makes
use of laboratory test conditions followed by a final verification based
on a field test.
\end{description}

It is not easy to produce a general recipe for the evaluation of a
certain group of applications. Specific requirements of a certain
application may lead to a different assessment methods, and at least to
different metrics and criteria.  Therefore, it is impossible to give
detailed examples of possible assessment projects. In general a robust
evaluation experiment should include experiments on various items of
user appreciation, system performance or technology.

Three examples are given:

Application oriented: The evaluation of a system using speech input and
speech output in a dialogue concept, such as used for a travel
information system, allows for the assessment of the total system (user
appreciation and system performance). Also the assessment of individual
modules may be relevant, consider the robustness of the recognizer for
telephone quality speech and the intelligibility of the \mbox{(text-to-)speech}
output system.  If the assessment is focussed on modules, many parameters
should be controlled.  However, for application oriented assessments one
can  use representative evaluation data where variables are
uncontrolled but more or less representative for the application.
In many cases a representative set of test data may be 
too big and therefore not realistic. But with a limited set of test data
valuable results can still be obtained. Also selective analysis of the results
(individual module responses) may produce diagnostic information on the
performance of the system.

Technology oriented: The present ARPA yearly competitive tests and the
present CEC-LRE SQALE project, in which various recognizers are
evaluated for multi-lingual use, are examples of technology oriented
projects. The present needs for performance under realistic conditions
require to cope with adverse conditions such as: telephone speech, noise
conditions, spontaneous speech, etc. Also text-to-speech systems for
many languages covered by the EU requires more attention. A competitive
project in which various systems are compared will push the technology.
It is also recommended that test data are made available to the
community.  Participation at a moment after such a project is launched
should be made possible.  It may be obvious that a general consensus on
this matter between partners and the CEC is required and that
dissemination is guaranteed. A possible example may be the results of
the SQALE project.

Combination of Spoken Language and Natural Language Projects: In general
SL and NLP use different assessment methods and scoring metrics.
Therefore it is not possible to give a realistic example of such a
combined project.  However, the use of common data bases for testing of
a combined system (e.g a translation system with speech input and/or
output, a spelling checker with voice control, or a information system
with an interpreter) may be a useful first step. Original text fragments
compared with distorted recognition outputs may be used for assessment
of the NLP module.

\subsection{Technology Assessment at Project Level}
\label{ta-proj}

\subsubsection{Progress Evaluation}

The role of project internal technology assessment is likely to be
primarily one of producing diagnostic information.  This
can be used to facilitate progress in system development, by identifying
gaps in the technology that need to be filled in order to improve system
performance.  It would serve to identify precise points in processing
where something is going wrong.
Internal technology assessment may or may not form part of
progress evaluation, which would in any case require a large element of
user-centered assessment.

To a large degree, internal technology assessment is a matter of internal
project policy.  It would be unwise to impose many general requirements.
First, the specifics of the technologies employed may
vary quite widely from project to project, with a concomitant variety
in the forms of assessment producing useful diagnostic information.
Second, some projects may aim to take speech or language processing modules
off the shelf and fit them into the system  with minimal adaptation.
Other projects may aim at more substantial refinement and development
of existing technologies. In the first case, it is not even clear that
technology assessment has any useful diagnostic role, and if it does it is
likely to take a different form from that in the second case.

Perhaps the most one can say is that projects should specify a progress
evaluation plan, and that where applicable this should include some
form of diagnostic technological assessment.  Technological assessment
is appropriate where a project envisages either (i) refining or
developing speech and/or language processing techniques, or (ii) adapting
data --- such as grammars, language models, lexicons --- for a particular
domain, task or language.  The form of the technological assessment can vary,
but in most cases one would expect some form of test-suite or test-corpus,
along with a specification of the results that should be obtained by processing
the test data.

\subsubsection{Relating Project Internal and Comparative Assessment}

There is an obvious tension between two claims that have been made
above. Namely, (a) that
the form  of project internal technology evaluation can vary from
project to project, and (b) that technology evaluation provides the core
of comparative assessment.  The second claim would seem to demand
uniformity between projects in technology evaluation, while the first
denies it.  This suggests that internal and comparative technology
assessment are very different kinds of animal. But
what one would like is for comparative technology assessment to build on
the back of project internal assessment.  If comparative assessment requires
a vast expenditure of extra effort, this will act as a strong disincentive
to participation.

Fortunately, the tension can be reduced. While internal
technology evaluation will tend to be more specific and detailed than
the comparative version, this is still compatible with there being an
a core of internal assessment that is common to a variety of projects.
In particular, different projects might very well use standard
annotations, such as labelled syntactic bracketings, basic predicate argument
structures, etc (as defined under Parseval and SemEval), to formulate
test data.  They may wish to extend the annotations, and produce other
kinds of data besides.  But if the standard components can be identified
and kept separable, then project internal test data may, with appropriate
extra effort, be re-usable for comparative purposes.

Another obvious point to make is that if an evolutionary system
development cycle is followed, then snapshots of a system
at different times can profitably be regarded, for the purposes
of comparative evaluation as different systems for the same domain,
task and user group.

Thus, in order to facilitate comparison and reusability of evaluation
data, a limited degree of standardisation is required for project
internal technology assessment.

\subsection{Input From the Speech and Language Community}

As part of the exercise of preparing this report, a questionnaire was designed
to sample opinion about technology assessment across the European Speech and
Language community. The questionnaire was not meant to be scientific, nor
exhaustive. It was intended to provoke response, which earlier trigger papers
has failed to do, and to see if there were very strong opinions in certain
directions.

More details concerning the questionnaire are presented in the Appendix. The
text of the questionnaire together with textual comments of the respondents
are to be found in section \ref{app:quest-text}; a tabulation of the results
of the questionnaire is to be found in section \ref{app:quest-results}. In
this section an attempt is made to summarise and to interpret, to some extent,
the responses to the questionnaire. Since this is always a treacherous
undertaking, readers are encouraged to consult the Appendix themselves for
more details.

The first question was intended to determine the attitude of researchers
towards two forms that technology assessment might take: project-internal and
comparative. The overwhelming majority of respondents expressed keenness to
conduct both sorts of technology assessment (some comments indicated
a confusion between project-internal technology assessment, and other forms of
project internal assessment, such as simple progress evaluation -- the
questionnaire may have failed to make this distinction sufficiently clearly).
There were worries expressed, however, about the cost of comparative
technology assessment and the possible effect that it might divert resources
away from other research and assessment activities.

The second, multipart question sought to explore the possible content of
project-internal technology assessment. The first subpart asked whether there
should be quantitative technology assessment of projects, in addition to any
(quantitative or non-quantitative) user validation that might be carried out.
Again, the overwhelming response was in favour of quantitative technology
assessment, independently of what user assessment might or might not take
place. The comments indicated that user assessment alone was not sufficient to
drive the technology or to promote reusability. The second subpart asked about
the level of granularity at which project-internal technology assessment
should take place, at the level of user-significant components or at finer
levels ? The majority preference was for assessment at finer levels. The final
subpart asked whether in project-internal assessment systems should be
assessed against test data that differed in various linguistic features (e.g.
language, domain etc) from the application for which it was designed. Response
was more cautious here. While the majority were in favour, comments indicated
that although it was clearly preferable to do this it might be too soon.

The third multipart question asked about implementing project-internal
technology assessment. The question aimed at determining whether researchers
were in favour of the evaluation exercise being largely run by the projects
themselves or whether it would be better for persons independent of the
project to be involved as evaluators and as evaluation data collectors. The
results indicated a clear preference for the evaluators to be project members
-- reasons cited were primarily the expense and bureaucracy the alternative
would entail. However, there was almost an even split for and against the
evaluation data collectors being members of projects.

The fourth multipart question aimed at determining views on the content of
comparative evaluation. The first subpart asked for preferences concerning an
integrated speech and language comparative evaluation exercise versus separate
speech and language evaluations. The split in responses was almost even,
with some people arguing for both. The second subpart asked about granularity
of comparative assessment (as with project-internal assessment), whether it
should be at the level of user-significant components only, or at finer levels
too. The bulk of the respondents supported finer levels of assessment, again
arguing that it would advance the technology; however fears were expressed
that development could be slowed by evaluation forcing a uniform modularity
across systems.

The next three subparts of the fourth question asked respondents to prioritise
user-significant system components they thought were candidates for
evaluation, sub-user-significant components suitable for evaluation, and
dimensions along which generality should be evaluated (and hence promoted).
For user-significant, written language components there was little consensus,
though document retrieval had perhaps the most support; for user-significant,
spoken language components, speech recognition was a clear winner with topic
spotting next. For sub-user-significant, written language components there
was again little consensus with part-of-speech tagging, morphological analysis,
named entity recognition and parsing attracting most support; for 
sub-user-significant,
spoken language components word lattice production, phone lattice production
and prosodic marking all attracted support but no
overwhelming preference was indicated. Finally, concerning dimensions along
which written language systems should generalised, domain was overwhelmingly
selected as most important, with language being placed second; for spoken
language, speaker-independence was clearly selected as the most important
generalisation required, while little consensus emerged about which other
features it was most important to generalise with respect to.

The fifth and final question asked about the implementation of comparative
assessment, attempting to see whether researchers supported a bottom-up
scenario, where the evaluation exercise(s) would evolve out of the actual pilot
applications funded under the first Language Engineering call in the Fourth
Framework, or whether they supported a top-down scenario where an evaluation
exercise would be defined in consultation with the community but 
independently of
particular funded pilots. There was a fairly even split here. However,
positions were hotly supported or rejected with comments ranging from
bottom-up is `just hopeless' to top-down `looks disastrous'. Readers are
encouraged to review the comments in section \ref{app:quest-text} under
Question 2.6. The fears about the bottom-up approach were that projects would
simply be too diverse to find common ground for evaluation and that such an
exercise could not be built on for future rounds when the projects might all
change; the fears about the top-down approach were that projects would be too
diverse for an external evaluation task to be imposed and that such an
exercise would turn into a bureaucratic nightmare.

To make a summary of the summary, one might tentatively draw the following 
broad
conclusions from the responses to questionnaire:
\begin{enumerate}
  \item technology assessment, as differentiated from user validation, is
        viewed as very important and needs to be carried out quantitatively, 
        both within
        projects and comparatively, and at a lower
        level than user-significant components -- it is seen as an important
        force in driving technology and encouraging reuse;
  \item projects themselves should carry out project internal technology
        assessment, though there may be a case for having evaluation data
        assembled by independent `experts';
  \item there is no consensus about how comparative assessment should be
        implemented;
  \item with the exception of speech recognition, there is no consensus
        about the most important evaluation exercises either for written
        or spoken language at either user-significant or user-transparent
        level;
  \item with the exception of domain for written language, there is no 
        consensus
        about the sort of generalisation that evaluation exercises
        ought to be encouraging.
\end{enumerate}

\section{Recommendations}

\subsection*{Overall Recommendation}

The ARPA evaluations have shown that there is much to be gained from
comparative evaluation, and the preceding discussion has indicated how
one can set up an evaluation exercise that can accommodate a variety
of different applications.  While technology evaluation must form the
core of the exercise, user-centered issues may readily be catered
for by paying due attention to the environmental aspects of technology
assessment.  Therefore we make the following general suggestions:

\begin{itemize}

\item In the initial stage a small scale comparative evaluation exercise
should be set up which is driven mainly bottom-up. It should have the
following features

\begin{itemize}

\item It should build as far as possible on materials gathered for
      project internal assessment plus other pre-existing resources
      where applicable.

\item Although it should start on a small scale, it is important that it
      has the flexibility to grow over time.

\item A flexible evaluation structure capable of accommodating a variety
      of tasks and systems, and with an emphasis on environmental /
      user-centered validation of technology measures should be
      employed.

\item The evaluation exercise should be open to LE sites not directly
      funded within the FP-4 LE programme, to encourage the spread of
      annotation standards, to bring in fresh ideas and experience, and
      to prevent the exercise becoming the preserve of a clique of
      established sites.

\end{itemize}

\item While steps should be taken to minimise costs, both for projects
participating in the evaluation and for central administration, it
should be recognised that such an evaluation exercise cannot come for
free.

Therefore, it may be desirable to provide limited extra resources to
those projects participating to cover additional efforts necessary for
participation.

However, participating projects should also gain from economies of scale
in setting up a core of common metrics and resources for evaluation.  In
due course, other projects within the FP-4 LE programme should also
benefit from these common resources and standards.

\item Provision would need to be made for a small group of people ---
independent of any specific project --- to oversee the collection,
refinement and dissemination of common evaluation material, as well as
the specification of evaluation architectures, standards and metrics.

This group could be either loosely organised in a support project or
they could be affiliated to some existing or to-be-established
organisation.

The coordinators would need to act in close cooperation with
participating projects in order to ensure that the exercise is built up
in such a way as to stay in touch with the technical and user needs of
the individual projects. The profile of expertise of this group should
reflect both the needs of technology assessment and of user
validation.

\end{itemize}

We add some more details to these recommendations below.

\subsection*{Project Internal Evaluation}

To a large degree, methods of project internal evaluation will be a
matter for negotiation between users and system developers within
individual projects.  However, there are a number of points that are not
only desirable for internal evaluation, but which would also facilitate
comparative evaluation:

\begin{itemize}

\item Projects should be encouraged to develop a well defined,
evolutionary evaluation strategy.

\item Projects should identify from the outset the kinds of evaluation
data that they will require, and the means by which this data is to be
acquired.  Early acquisition (and use) of evaluation data should be
encouraged.

\item Projects should provide functional specifications that not only
identify tasks and sub-tasks, but also the environmental attributes
pertinent to those tasks.

\item Wherever possible, projects should be encouraged to make their
internal evaluation strategies, test data, user profiles etc. available
to a wider community, even if they are not willing to participate in a
comparative evaluation exercise.

\end{itemize}

Section~\ref{geert} amplifies on some of these points, and suggests some
practices for project internal evaluation.

\subsection*{Initiating Comparative Evaluation}

The comparative evaluation exercise needs to be set up in a way that
is part top-down and part bottom-up. A number of steps need to be taken 
to initiate the exercise

\begin{enumerate}

\item A small number of projects should be selected to take part in the
exercise. The projects must be willing to participate, and they must all
be geared to tasks that have a substantial degree of overlap.  Clear
agreement must be reached at the outset about public accessibility of
the evaluation data that projects will be supplying.

\item A braided task structure should be identified to accommodate the
participating projects.  This may involve further refinement of the
functional specifications employed by some of the projects.

\item Natural evaluation points in the braided structure should be
identified. Under realistic assumptions, their number will not exceed
half a dozen such points.

\item Appropriate corpus annotations for the selected evaluation points
should be specified.  The specification should be carried out in
consultation with participating projects, but where possible standard
annotations (word strings, morphosyntactic tags, labelled syntactic
bracketings, predicate-argument structure, co-reference relations, scope
relations) should be used.  It is important that the annotations allow
for the measurement of both accuracy and depth of analysis.

\item Projects whose task structure includes a particular evaluation
point are expected to employ the annotation relevant to that point as
part of their internal assessment regime. (They may of course also apply
additional assessment measures).

\item Tools should be devised for (a) producing (target) annotations,
and (b) measuring actual against target annotations.  One should aim to
exploit existing tools where possible, and otherwise distribute
development effort over different projects and the evaluation
coordinators.

\item Using these tools, individual projects should produce annotated
answer data from their own evaluation corpora.  These should be passed
on to the evaluation coordinators.  It is then the job of the
coordinators to assess the suitability of the material as a basis for
common evaluation data. Producing common evaluation data will typically
involve the coordinators, perhaps in conjunction with other
participating projects, refining and extending the range of annotations
on the material delivered.

\item Individual projects should make available as much additional
linguistic and non-linguistic data pertinent to evaluation as possible.
Presenting this data in a form accessible to other projects will be
problematic unless some kind of interchange format is agreed.

\end{enumerate}

Having obtained this initial infrastructure it is then necessary to attempt
to validate the chosen common annotations and evaluation metrics.
This involves correlating metric scores with the adequacy or inadequacy
of task performance under the different environmental attributes imposed
by different systems.  This correlation would be greatly assisted if
systems were assessed at different stages of development.  This would allow
comparison of technology measures and system adequacy under relatively
fixed environmental constraints.  It should be borne in mind that this
form of validation may well show that some of the originally selected
technology metrics and annotations are of limited or zero validity.

\subsection*{Central Coordination and Resources}

Even a small comparative evaluation exercise cannot be expected to be
self-running and self-regulating.  Some central coordinating
organisation, independent of the participating projects, is needed.  The
function of this coordinating organisation would be to negotiate with
the projects participating in the comparative exercise, to collect,
administer and disseminate evaluation material, and to survey and
synchronise the evaluation exercise.

The central coordination in the initial stage could be in the
responsibility of some support project, e.g., the European Linguistic
Resource Agency (ELRA) or a comparable organisation involving a group of
independent experts. The profile of these experts should include
industrial and academic expertise in the areas of evaluation and
standardisation, language technology, linguistic and software
engineering, as well as knowledge of sectors and application areas where
user centered assessment is concerned.

Where possible, the evaluation exercise should build on pre-existing
materials, such as data (and expertise) from the various ARPA
evaluations, the Penn tree-bank, various national European exercises
(e.g., the French GRACE exercise), results of LRE projects such as
MULTEXT and TSNLP, and so forth. For the common annotations employed, it
could be desirable to draw on the annotation schemes devised under
Parseval and SemEval, where appropriate.

\appendix

\section{Questionnaire}

\subsection{Text of the Questionnaire}
\label{app:quest-text}

The text of the questionnaire follows in {\tt teletype font}.
After each question selected comments from respondents
have been included in {\em italic font}, prefixed by
the respondent's numeric response, if any, to the question.

\footnotesize
\begin{verbatim}
                QUESTIONNAIRE ON ASSESSMENT AND EVALUATION
                ==========================================

                       IN LANGUAGE ENGINEERING
                       =======================

1.0 Background
==============

The EC has set up a Study Group on Assessment and Evaluation (A & E) in
Language Engineering (LE) with a view to establishing productive ways 
forward within Framework IV (and beyond) in the area of assessing and 
evaluating both particular systems and underlying LE technologies. Your 
views on this subject are sought.

As you will no doubt be aware, the ARPA-sponsored evaluation programmes 
in the United States (CSR, ATIS, MUC, TREC, etc.) have become extremely 
influential in shaping the direction of LE research there. The Commission
is now considering whether there should be a set of European actions in 
A & E and if so, what form they should take. To that end the Study Group 
has been set up and rapporteurs for a number of subgroups appointed. 
Several trigger papers have been prepared which some of you may have seen
(these are available by non-anonymous ftp from cl-ftp.dfki.uni-sb.de with
username 'assessment' and password 'tnemssessa'). However, these have not
led to much response as yet. In order to provoke more response a 
questionnaire has been developed and is attached below. I would be most 
obliged if you could complete the questionnaire and return it to me as 
soon as possible. Responses will feed into the production of Commission 
policy concerning requirements to be placed on individual projects 
concerning A & E and also concerning specific initiatives in this area. 
Due to time constraints on the production of position papers, responses 
must be received no later than March 10 to be taken fully into account.

In the current Telematics Language Engineering call, pilot applications 
(which form the bulk of the projects under Framework IV) are being 
assessed along the two broad dimensions of:
1. user validation; and
2. technology assessment and system evaluation (TA & SE).

This questionnaire pertains chiefly to the latter dimension.

2.0 Questionnaire
=================

2.1 Introduction and Terminology
================================

Terms later used in the specific sense introduced here are capitalised.

An LE APPLICATION SYSTEM or just LE SYSTEM is a set of software components
constructed to permit a user to carry out some language-related function 
in a specific real-world environment.

LE systems and system components may be described in terms of:
1)  the LINGUISTIC FUNCTION they carry out (e.g. translation, speech
    recognition, parsing, summarising, coreference resolution) -- 
    components which have some broadly agreed linguistic function will be
    termed LINGUISTIC PROCESSING (LP) COMPONENTS (so, e.g., a parser is a
    linguistic processing component whereas a binary search routine is not)
2)  LINGUISTIC FEATURES of the input or output data they operate on or 
    produce (e.g. language (French,English), subject area or domain 
    (weather reports, financial newswire stories), text type (newswire 
    stories, technical manuals), text length (paragraph,article), 
    spontaneity (read speech vs. spontaneous speech), channel conditions 
    (telephone vs. wide-band), and so  on)
3)  whether or not the inputs and outputs are objects which are of 
    functional significance to the user of the system -- systems or 
    components whose inputs and whose outputs are of such significance 
    will be termed USER-SIGNIFICANT (e.g. a tokeniser whose input is an
    English newswire story but whose output is a sequence of pointer pairs
    indicating token start-end positions is not a user-significant 
    component; a system with the same input whose output is the same 
    newswire story in French is user-significant).

To evaluate a system and the technology it embodies, a decision must be made
about the level of:
1)  GRANULARITY at which it will be evaluated (e.g. at the level of
    user-significant components only, or at some level of LP components
    which are sub-user-significant)
2)  GENERALITY at which it will be evaluated (e.g. how much do we vary the
    linguistic features of the input/output data we provide/expect relative
    to the features of the data in the intended application; e.g do
    we evaluate the system against data from different languages, different
    domains etc.).
These two dimensions are orthogonal: the decision to evaluate at a high 
level of granularity is independent of the decision about which linguistic 
features of the input/output data are to be generalised, if any, and to what
extent.

By PROJECT INTERNAL TA & SE is meant assessment measures devised for and
applied to individual projects, with no attempt made at isolating and
comparing technologies or LP components across projects.

By COMPARATIVE TA & SE is meant the complement: assessment measures which
attempt to isolate and compare LP components which recur in different 
systems.

Example
=======

To make these distinctions clearer consider this example. MET is a
hypothetical LE system that translates spoken English reports composed by 
the British Metereological office into French text suitable for reading by 
French announcers on local radio stations in Britanny. The linguistic 
functions of the overall system, and those which define user-significant 
components, may be described as transcription and translation. But we may 
suppose the system contains linguistic processing subcomponents which are 
not user-significant and which accomplish such functions as word lattice 
interpretation, part-of-speech tagging, parsing, English-French syntax tree
transformation, etc. Relevant linguistic features of the input data include:
English language, single speaker, read speech, limited domain, short 
passages, formulaic style; relevant linguistic features of the output are 
French language, formulaic style, limited domain.

To assess the system we may want only to see how the user-significant
components behave on unseen data whose linguistic features match those of 
the intended application; to assess the technology we may want to evaluate 
both at a lower level of granularity (so e.g. the word lattice 
interpretation component may be of interest) and by generalising the values
of the linguistic features of the input/output data (so, e.g. seeing if the 
techniques employed may be generalised other languages, other sorts of brief
report, etc.).


2.2 General Form of Assessment
==============================

  1)  PROJECT INTERNAL TA & SE

      What is your attitude towards conducting project internal TA & SE ?

      1. Would not want to conduct it at all
      2. Would conduct it, but only if necessary
      3. Would be willing to conduct it
      4. Would be keen to conduct it

      Response:       (1 - 4)
      Comment:

\end{verbatim}
{\em

4: It is essential at the technology level to know where we are


4: Project internal evaluation is absolutely necessary.
               Without systematic evaluation it is impossible to measure
               progress.
       Maybe you should have an option 5. I see quantitative evaluation as
         indispensible, even if it is very time consuming

3:      Presumably this only has a point if you do it several times
        within a project, as a measure of progress (well, change at
        least). For an applied project, it ought to be part of the basic
        research methodology. For more theoretical work its more
        problematic since you might not know what to measure, and it
        may change as the project progresses.

4:   [A] project that does not involve some form of
      at least progress evaluation would not be properly managed

2: Any project has to have some way of deciding how well it
        is doing.  I'm assuming you mean something that has been regimented
        from outside.
}

\begin{verbatim}
  2)  COMPARATIVE TA & SE ?

      What is your attitude towards participation in comparative TA & SE ?

      1. Would not want to participate at all
      2. Would participate, but only if necessary
      3. Would be willing to participate
      4. Would be keen to participate

      Response:       (1 - 4)
      Comment:

\end{verbatim}
{\em

4: It is more difficult (therefore be very careful in
defining reasonable goals), but necessary

4:I'm more in favour of comparative TA/SE if broadened and
funded to include non-LE specific projects.  If this doesn't happen
then from among only the EU-funded projects there would be insufficient
overlap so no point in trying to compare heterogeneous systems.

4: We already actively participate in common comparative
               evaluations. This is an important means for us to compare
               our work with what others do, to learn from their techniques,
               and to improve our research and system.

4:      Properly resourced participation would be a useful discipline.
        Not convinced how useful it would be, but that's not what you
        asked. In common (I suspect) with much of the community,
        however, I have much more interest in participating than in
        developing the substantial framework that would be required to
        allow participation - maybe there's another community out there
        who might be interested in doing that kind of thing?

4:  This assumes that
    \begin{enumerate}
      \item comparative assessment is not going to divert substantial
         resources from internal assessment and development;
      \item that comparative assessment is likely to bring about
         project internal improvements, in the same way that
         internal evaluation should;
      \item a sensible framework can be found for conducting comparative
         assessment across a variety of projects covering different
         tasks and application domains; in particular, methods should
         not be biassed by the adoption of one particular linguistic
         theory or another.
    \end{enumerate}

2:  I'm not keen, but if all money comes with this sort of
        strings attached, I haven't got much choice.

4:  Concern that too much research time would be
    directed towards setting up evaluations (but this had good results
    in the ARPA community)

}
\begin{verbatim}

2.3 Content of Project Internal TA & SE
=======================================

  1)  USER VALIDATION & TECHNOLOGY ASSESSMENT

      Should projects

      1.  rely on user validation only (i.e. whatever validation the users
          deem sufficient) ?
      2.  rely on user validation and additionally, if not specified in user
          validation, produce test data (input/output pairings) and carry 
          out tests which allow quantitative assessment of performance ?

      Response:       (1 - 2)
      Comment:

\end{verbatim}
{\em

2: Assessment should be as quantitative as possible. Very
important here is the number and representability of users tested on and
that each test should be made on new data with frequency of about once a year.

2: Quantified numbers tell the truth, or appear to.

2: qualitative assessment is not enough. (only user validation would
               require a huge number of users of all types to have a 
               statistically
               meaningful result. It is well known that most users do not know
               what they want or need)

2:      Hey, you're changing the rules here. In your intro you
        specifically separated User Validation (UV) from TA \& SE, and
        now up pops UV as part of TA \& SE. But it also changes the rules
        in another way, because I don't see UV as project internal - if
        two projects attempt to do the same thing, then their UV results
        can be compared. Once you can do that, the considerations
        change.
        I actually think it may be dangerous to place too much emphasis
        on UV, because depending on users' existing preconceptions
        introduces a huge inertia into technological development. So
        even if one wants to steer projects into user-oriented
        directions (as the EC clearly does), one needs to allow for
        assessment independent of any existing user base and encourage a
        degree of technology-pull in users.

2:  User validation alone is unlikely to provide sufficient
      diagnostic leverage for technological development.  Test data
      at least provides some moves in this direction.

2:  User validation alone won't promote generality and
        reusability of components.

}

\begin{verbatim}

  2)  GRANULARITY

      Supposing quantitative assessment against test data, at what level
      of granularity should linguistic processing components be evaluated ?

      1. at the level of user-significant components only
      2. at finer levels as well

      Response:       (1 - 2)
      Comment:

\end{verbatim}
{\em

2: User validity is very important but often very difficult to
define quantitatively. Therefore technological assessment levels are
undetournable.

2: Finer evaluation is necessary for system development

1,2:    So 2 was the right answer to Q1 eh? Well you've broken the
        tidy model now so there's no clear answer. For semi-objective UV
        it only makes to look a user-significant components. As a
        project internal measure you would probably be well advised to
        look at finer levels too, although you might sometimes feel you
        didn't need to.

2: Again, for the purposes of diagnostic leverage.  But clearly,
      the measures would not be of interest to users.

2:  User validation alone won't promote generality and
        reusability of components.

1: Difficult to meaningfully evaluate at finer levels
}

\begin{verbatim}

  3)  GENERALITY

      Supposing quantitative assessment at what level of generality should
      components be evaluated ?

      1.  only on unseen input-output pairings whose linguistic features 
          match those of the project application
      2.  on input-output pairings whose linguistic features differ from
          those of the project application,  as well as on those whose
          linguistic features match those of the project application

      Response:       (1 - 2)
      Comment:

\end{verbatim}
{\em

 1: Answer 2 might be interesting in at least 2-3 years time. We
could try to push more and ask for multilinguality from now on (like in SQALE)
and (more difficult because it needs more data) on different domains, but
we might encounter the risk of frightening people.

1: While I agree that 2 would be nice, I do not think that we
               are there. Most of the results would be pretty meaningless.
               When possible, 2 is clearly preferable.

1,2:    This is a silly question. You will achieve different things
        depending which you do. It just depends on what you want to know
        about your system. True blue user-orientation would argue that 1
        was enough. More generally-oriented work would do well do think
        about 2. Some recent EC calls seem to want to do both, which
        isn't obviously sensible - that is, doing user-oriented work and
        then looking to see how general it turns out to be is not a good
        way to make useful progress.

2: This answer needs to be qualified.  What it would be useful
      to measure is how readily a system  can be ported from one domain
      to another.  This reflects a system's maintainability, portability
      and flexibility.  However, provision should be made for spending
      a certain amount of time porting between domains, and not just
      doing it straight off

2:  User validation alone won't promote generality and
        reusability of components.

2: Case 2 can only give more information compared with case 1.
}

\begin{verbatim}

2.4 Implementation of Project Internal TA & SE
==============================================

  1)  EVALUATORS

      Should tests be carried out

      1. internally by project participants (developers and users) ?
      2. externally by independent project evaluators ?

      Response:       (1 - 2)
      Comment:
\end{verbatim}
{\em

1: Answer 2 is interesting for projects that are very near
to products (within 6 months on the market)

1:makes it seem too forced.  Projects should want to do this.

1: we need to have confidence in the participants now.
               In our area (spoken) it would be extremely difficult to have
               external evaluators. Maybe in the written areas it
               is easier.
1:      Again, this depends what you want. But this time I'll express an
        opinion about what I want and its 1. I think 2 is overkill given
        the fairly limited objective utility of project-internal TA \&
        SE. It should just be part of the research team's methodology
        and they should do it primarily for project-management purposes.

        (If anyone thinks there's a problem about integrity and trust of
        research teams I don't think 2 will actually make the teams
        better, though it might make the administrators feel better.)

1: Assuming that the intended users count as project internal.
      External evaluators are likely to be a bureaucratic nightmare, and
      expensive

-: Depends on cases again. 2 will involve a lot of organising
        and doesn't make sense for every project.

-: If the infrastructure was in places then (2) would be
         preferable, but I'm not convinced that it is the best use of
         resources to set up such an infrastructure effectively from scratch.

}

\begin{verbatim}

  2)  EVALUATION DATA COLLECTION

      Should test data be compiled

      1. internally by project participants (developers and users) ?
      2. externally by independent specialists ?

      Response:       (1 - 2)
      Comment:

\end{verbatim}
{\em

2: That would be nice if we could find independent specialists.

1: makes it seem too forced.  Projects should want to do this.

2: but these specialists must do so in close coordination
               with the partners to ensure compatibility/applicability

1:      See previous comment. But there's the additional proviso I
        mentioned above that its not obvious that the kind of people who
        typically propose the research are the same as the kind of
        people who want to construct test data - maybe projects should
        be encouraged to include 'independent specialists' in their
        consortia.... (if such specialists really exist?)
1:  It ought to be cheaper to collect this stuff internally.
      But not for data going beyond the project application, as in 2.3.3
      above.  Here it would be useful if some other project could collect
      the data internally, and make it available.

-: Depends on cases again. 2 will involve a lot of organising
        and doesn't make sense for every project.
}

\begin{verbatim}

2.5 Content of Comparative TA & SE
==================================

  1)  INTEGRATED VS NON-INTEGRATED SPEECH AND LANGUAGE EVALUATION

      The Commission should aim to develop

      1.  an integrated speech and language evaluation.
      2.  independent evaluations for spoken and written language 
          engineering.

      Response:       (1 - 2)
      Comment:
\end{verbatim}
{\em

1 for technology pushed projects, 2 for projects where user
validation is required (application oriented projects).

2:insufficient overlap between the two to date ... too early

Both: it really depends on the application. when possible the
               paradigm should be integrated, though many issues are very
               different. However, spoken and written language have a lot
               to learn from each other.

2:      I think there will always be sharp distinctions between the
        user-significant bits of speech and text systems, even if
        internal modules can be shared. I also think there will be for a
        long time some tasks which are specifically speech and others
        that are specifically text (plus some which might be both). So
        it would be wasting effort to try and force and integrated view
        in this way - the actual applications would still end up being
        polarised one way or the other.

        But actually we should be considering a three-way distinction:
        speech, text, and common ('language'?) - most of the latter
        not being user-significant. Then there's handwriting as well....

1: However, speech only and language only systems should
      be able to fit in as well

-:  Not a sensible pair of alternatives.  A single integrated
        form of evaluation to cover all possible projects is impossible.
        It doesn't follow that speech should be split off from the rest of
        language.

-:    This depends on the application but in general you need
                     both

1: Option 1 is more risky in terms of complexity and
                possible biases.

-: Integration is over strong; there ought to be some connection
               but independent cases are also desirable.

}

\begin{verbatim}

  2)  GRANULARITY

      At what level of granularity should linguistic processing components 
      be evaluated ?

      1. at the level of user-significant components only
      2. at finer levels as well

      Response:       (1 - 2)
      Comment:
\end{verbatim}
{\em

2: I am of the opinion technology is still to be improved, which
means the finer levels are of high importance.

2:   Finer evaluation is necessary for system development

1:  I know MUC-6 has gone for 2, but my tendency is to think that
        you can't pin down internal structure sharply enough to be
        useful without compromising development. You might think you can
        (as in Hobb's generic MUC system), but in practice aligning real
        modules with these ideals doesn't really work out.
        I'm all for modularity, and component re-use, for components
        that are not the focus of development in a project. But
        evaluation requires modularity to be enforced everywhere, and
        that will just block innovation.

2: We need to bear in mind that finer granularity may differ
      substantially between systems, so that in may cases comparative
      evaluation may not be possible. But where it is, it is likely to
      provide useful diagnostic information
}

\begin{verbatim}


  3)  CANDIDATE USER-SIGNIFICANT COMPONENTS FOR EVALUATION

      Given the following user-significant linguistic functions, select 
      those in which you would be willing to participate in a comparative
      evaluation and prioritise them (do this by writing your priority 
      number after the task; feel free to extend the list):


      For written language:
                                                            Your priority
      1. translation                                            ?
      2. summarisation                                          ?
      3. document retrieval by topic                            ?
      4. template filling or information extraction             ?
      5. question answering                                     ?
      6. generation                                             ?

      For spoken language:

      1. recognition                                            ?
      2. topic spotting                                         ?
      3. generation                                             ?
\end{verbatim}
{\em

I don't know what 'willing' is supposed to mean here - this is a
very resource-dependent thing.
}

\begin{verbatim}

  4)  CANDIDATE SUB-USER-SIGNIFICANT COMPONENTS FOR EVALUATION

      Given the following language processing functions which are probably
      below the level of user-significance, select those in which you would
      be willing to participate in a comparative  evaluation and prioritise
      them (do this by writing your priority number after the task; feel 
      free to extend the list or to specify more precisely the form the 
      analysis would need to take for you to be interested):

      For written language:
                                                            Your priority
      1. named entity recognition (company names,               ?
            locations, personal names)
      2. part-of-speech tagging                                 ?
      3. morphological analysis                                 ?
      4. parsing                                                ?
      5. coreference resolution                                 ?
      6. word sense identification                              ?
      7. predicate-argument structure identification            ?

      For spoken language:
      1. word lattice production                                ?
      2. phone lattice production                               ?
      3. prosodic marking                                       ?
\end{verbatim}
{\em

many spoken language systems wont produce any of these.
}

\begin{verbatim}

  5)  GENERALITY

      With respect to which linguistic features of the input/output should
      a comparative evaluation attempt to encourage generality ?
      (write your priority number after the feature; feel free to extend the
      list or to replicate it for different linguistic functions -- e.g.
      language independence may be more important for advancing translation
      technology than for advancing summarisation research):

      For written language:
      1. language                                               ?
      2. subject area or domain                                 ?
      3. text type                                              ?
      4. text length                                            ?

      For spoken language:
      1. speaker-dependence                                     ?
      2. spontaneity                                            ?
      3. channel bandwidth                                      ?
      4. native-non-native speakers                             ?
\end{verbatim}
{\em

}

\begin{verbatim}


2.6 Implementation of Comparative TA & SE
=========================================

Two possible scenarios for developing an evaluation programme are the
following.

Bottom-up scenario: pilot applications funded in the Telematics LE 1st call
(closing March 15) are required, where appropriate, to participate in the
provision of data and the definition of one or more comparative evaluation
exercises perhaps with assistance from external specialists (these exercises
might be designed around project `clusters'). A project is funded under the
`resources' heading in the 2nd call (to be announced Sept 95) to coordinate
the definition of the evaluation exercises and the preparation of test data
and scoring software. Selected pilot applications receive funding, where
appropriate, under the final call (Sept 96) to participate evaluation
exercises. In this scenario the type of evaluation evolves out of actual
funded pilot applications.

Top-down scenario: an evaluation exercise is defined through consultation 
with the community, a project is funded to coordinate it and to produce 
test data, and several projects (a very small number) are funded exclusively
to take part in it. This would be done in the 2nd and final calls. In this 
scenario the evaluation is initially designed separately from actual LE 
applications, but the aim is to produce a framework which could expand to 
attract a range of participants involved in application work.


  1)  COMPARATIVE EVALUATION IMPLEMENTATION SCENARIOS

      Which of the two scenarios do you believe is the best approach for
      establishing a European comparative evaluation exercise ?

      1. the bottom-up scenario ?
      2. the top-down scenario ?

      Response:       (1 - 2)
      Comment:
\end{verbatim}

{\em
 2: The top-down scenario is very important for technology
assessment, but user validation plays also a role and some selected projects
should receive funding to make a user validation, which is totally different
in spirit to technology assessment and very project or prototype dependent.

1: ... bottom up ... there will be too few LE projects to impose, a
priori, an evaluation scenario.  TREC works 'cos there are 57 groups
taking part in TREC4 and they all do the same ad hoc a/o routing
experiments, i.e. homogeneous. LE projects will be heterogeneous.

2:  I do not believe that 1 will work. What does "are required,
                where appropriate,  to participate ..." I am afraid that this
                will not lead anywhere. Also, the type of evaluation will
                likely be only valid for actual pilot applications, and may
                not extend to future projects.
                2 is more oriented to technology push, and should provide
                a more general framework. However, for it to work the
                whole evaluation process a learning experience for
                all participants, providing detailed information exchange.

2:      I don't think either of theses is an effective way to achieve
        this goal, but 2 comes closest. 1 is just hopeless - the idea
        that you will be able to induce useful evaluation scenarios from
        the projects that happen to get through the 1st call is not
        realistic - there won't be enough of them, and they will be too
        diverse (almost by definition - LE won't want to fund many
        groups doing the same things, unless it ALREADY has a reason
        to do so.)
        But the description of 2 suggests it will be hopelessly
        underresourced and if so it will be a waste of effort too - just
        one little evaluation will result, plus perhaps a methodology
        for doing more. But is there a reason to believe that will tell
        us more than we already know from the American experience?

1: It would be nice if some top-down scenario could be developed,
      with several groups working on exactly the same problems.  However, given
      the structure of the LE call, this clearly won't happen, or if it did
      it would exclude most of the LE projects.  A bottom-up approach seems
      the only way forward.

1: 1 is not great, but 2 looks disastrous.

1: Top-down sounds like mega-machine bureaucratic nightmare...

-: I do not believe one should make a priori choices here
               for example a great deal depends on what is actually
               submitted as pilots.
}

\begin{verbatim}


2.7 Other Comments
==================

    If you have any other comments you would like to make about the way you
    think TA & SE should be carried out in the EC, or about aspects of this
    questionnaire (omissions or commissions) then please make them here.

    General comments:
\end{verbatim}

{\em

-- I have some difficulty to understand in what role I (or
actually anyone else approached) should respond to your questions:
\begin{itemize}
    \item as a member of the study group
    \item as somebody with specific feelings about this whole field of LE 
         assessment
    \item or as an actual or imaginary partner in ongoing or future LE projects
\end{itemize}

-- It is for me much too complex to have the pretension to make a user 
validation
assessment and a technological evaluation at the same time: emphasize should be
made on one or the other. User validation is very important for near to
product projects: it should also include technological evaluation to
quantify with different objective measures the user validation process.
Technology evaluation is important to push the technology further, for
example in speech recognition to push to spontaneity or multilingualism or to
larger domains e.g. general business letter dictation. Technology development
project are important in order to have a good basis for future more ambitious
application oriented projects.

-- I think that the most important issue is that evaluation be taken
seriously in order to iteratively improve technology and applications.

-- My answers are very much those of someone interested in developing,
    rather than using, written language technology.  From that point of
    view, evaluation is a waste of time unless it can provide diagnostic
    information pointing out gaps in the technology.  It would be nice
    if it could also be established that filling these technological gaps
    also leads to better systems from a user point of view.  Or at any
    rate, which (if any) technological gaps detract from user
    adequacy.

-- This all looks dangerously bureaucracy-driven.  Administrators will find
it convenient to have numbers that can be put in rank order.  I think there
is a great risk of stifling new work that doesn't fit the pattern.  An
occasional open competition of the DARPA sort sounds like a fine idea, but
making it a model for all work does not.
There is also too much scope for creating a class of professional evaluators,
who will be people that couldn't make it in research.  There are already
some groups like this around.  The first item on their agenda is, naturally
enough, to keep themselves in business, which is not the same as promoting
good work.
The background material above talks about "independent experts", but real ones
will be extremely hard to come by.  For one thing, serious experts have their
own work to do.  For another, in a small community like this, people with
enough expertise to be of use are very likely to be either colleagues or
competitors - or a bit of both - of the people being evaluated.

--
While interested in trying to do it, I have considerable difficulties
in trying to answer the questions in your questionnaire for the following
reasons:

\begin{enumerate}
\item you set up false dichotomies, where one actually has a gradation
eg for 2.2, one end is assessing just one individual application in its
own right and the other is comparing systems regardless of contexts of
use . Clearly some mix is required.

\item its not obvious whether your questions should be answered from a
realistic or an idealistic point of view about what sort of evaluation 
programme
one might go for.

\item there's too much of a top-down flavour about things, as if one is
saying, OK lets evaluate (for its own sake), so whats the best way of
doing things. A more appropriate view would be to have some desiderata
and consider various possible evaluation suggestions or scenarios against
these: this seems to me the only sensible way of getting a fix
on what LP systems/components are of value in relation to the
conditions that make them of value.
\end{enumerate}
}

\normalsize

\subsection{Tabulation of Responses}
\label{app:quest-results}

The total number of responses to the questionnaire was 11 out of 35.
The following tables are an attempt to summarise the results in
tabular/numeric form. Because of the recalcitrance of the respondents,
who frequently refused to play by the rules (this awkwardness was
encouraged) these numeric summaries should not be taken too
seriously, or interpreted too literally. In many cases the
comments attached to the responses are of far more importance.

\subsubsection{Numeric choice questions}

Not all respondents replied to all questions. In some cases respondents
indicated that they thought there was no sensible answer to the
question, or that they had no opinion, or that more than one answer
was appropriate depending on various factors.

\begin{tabular}{|lll|c|c|c|c|} \hline
\multicolumn{3}{|c|}{Question} & \multicolumn{4}{c|}{Response} \\
                            \cline{4-7}
&&                                        & 1     & 2     & 3     & 4 \\ \hline
2.2 &&General Form of Assessment          &       &       &       & \\
    &1 &Project Internal TA \& SE         &       & 1      & 3     & 6  \\
    &2&Comparative TA \& SE               &       & 1      &  1 & 8 \\ \hline
2.3 &&Content of Project Internal TA \& SE&       &       &       & \\
    &1 &User Validation \& Technology Assessment &    & 10   &       & \\
    &2 &Granularity                       &  2    & 7      &       & \\
    &3 &Generality                        &  2    & 5     &       & \\ \hline
2.4 &&Implementation of Project Internal TA \& SE&    &   &       & \\
    &1 &Evaluators                        &  7    & 1      &       & \\
    &2 &Evaluation Data Collection        &  4    & 5      &       & \\ \hline
2.5 &&Content of Comparative TA \& SE     &       &       &       & \\
    &1 &Integrated vs Non-Integrated      &       &       &       &  \\
    && $\;$Speech and Language Evaluation &  5    & 4      &       & \\
    &2& Granularity                       &  3    & 7     &       & \\ \hline
2.6 && Implementation of Comparative TA \& SE &  &       &       & \\
    &1& Comparative Evaluation            &       &       &       & \\
        && $\;$Implementation Scenarios   &  4    &  5     &       & \\ \hline
\end{tabular}

\subsubsection{Ranked priority questions}

These tables show how many selected which option in which priority position.
I.e., for each item the number of persons selecting that item at a given 
priority
ranking is shown. Only top five priorities are listed.

Not all respondents answered all questions. Some did not use numeric ranking,
but instead introduced terms such as `important', `very important' etc.
sometimes putting several items into the same category. I have mapped these
onto numeric ranking as follows. Whatever category ordering was adopted has
been mapped onto the numeric ranking from 1 down in sequence (e.g. `very
important', `important', `fairly important' are mapped onto 1, 2, 3
respectively.). If several items are listed at the same ranking level that
level's count of 1 is distributed proportionally over the items (e.g. if
parsing and tagging are both `very important' the count in position 1 for each
of parsing and tagging goes up by .5). In at least one case one item
was deemed 'uninteresting'; this was recorded by introducing a
`-' column heading, to indicate a negative priority.

There were three ranked priority questions in the questionnaire all part of
section 2.5 Content of Comparative TA \& SE. Their purpose was to ascertain
crudely what sort of activity people would like to see in comparative
evaluation exercises.

\begin{tabular}{|lll|c|c|c|c|c|} \hline
\multicolumn{3}{|l|}{2.5.3 Candidate User-Significant} & 
\multicolumn{5}{c|}{Ranking} \\
                            \cline{4-8}
&&\multicolumn{1}{l|}{Components for Evaluation}
                                            & 1 & 2 & 3 & 4 & 5\\ \hline \hline
& \multicolumn{2}{l|}{Written Language}     &      &      &      &    & \\
&\hspace*{.5cm}& translation                & 1.5 & &  &       & 1\\ \hline
&& summarisation                            &  &  1     &   &     & \\ \hline
&& document retrieval by topic              & 2     &  2     &  & & \\ \hline
&& translingual retrieval                   &       &  1    &   & & \\ \hline
&& template filling/information extraction  &  1     &      &  2 & & \\ \hline
&& question answering                       & 1.5   &  &   &       & \\ \hline
&& generation                               &       &   1    &  & 1 & \\ \hline
& \multicolumn{2}{l|}{Spoken Language}      &      &      &      &    & \\
&& recognition                              &   6   &  &   &       & \\ \hline
&& topic spotting                           &     &  3    &  &   & 1 \\ \hline
&& generation                               &   1   &  1    & 1  &  & \\ \hline
&& language identification                  &   &  &  1     &       & \\ \hline
&& speaker  identification                  &  &  &       &  1     & \\ \hline
&& document retrieval by topic              &  &  1 &      &       & \\ \hline
&& template filling/information extraction  &  &  &       &       & \\ \hline
&\hspace*{.5cm}& translation                &  &  &       &       & \\ \hline
\end{tabular}

\vspace*{.5cm}

\begin{tabular}{|lll|c|c|c|c|c|c|} \hline
\multicolumn{3}{|l|}{2.5.4 Candidate Sub-User-Significant} & 
\multicolumn{6}{c|}{Ranking} \\
                            \cline{4-9}
&&\multicolumn{1}{l|}{Components for Evaluation}
                               & 1 & 2  & 3     & 4   & 5 & - \\ \hline \hline
& \multicolumn{2}{l|}{Written Language}     &   &       &       &     & & \\
&\hspace*{.5cm}& named entity recognition   &  .3     & 2 &  &  &   &\\ \hline
&& part-of-speech tagging                   &  2     &  1  & &  &   &\\ \hline
&& morphological analysis                  &  1.3    &     & &  &   &\\ \hline
&& parsing                                  &  .25   &  .5 & 1 & &  &\\ \hline
&& coreference resolution                   &  .25   &     & & & 1 &\\ \hline
&& word sense identification                &  .55   &     & & 1 & &\\ \hline
&& predicate-argument structure             &  .25   &  .5 & &   &  &\\ \hline
& \multicolumn{2}{l|}{Spoken Language}      &       &  &      &    & & \\
&& word lattice production                  &   3   & 1 & &   & & \\ \hline
&& phone lattice production                 &       &  3 & &  & & \\ \hline
&& prosodic marking                         &   2   &  & 2 &  &  &1\\ \hline
&& first sentence production                &   1   &  &   &  & & \\ \hline
&& pronunciation lexicons                   &       &  &   & 1 & & \\ \hline
&& vocabulary lists                        &       &  & 1  &  & & \\ \hline
&& semantic frame representation            &       & &  & &1 & \\ \hline
\end{tabular}


\begin{tabular}{|lll|c|c|c|c|c|} \hline
\multicolumn{3}{|l|}{2.5.5 Linguistic Feature} & 
  \multicolumn{5}{c|}{Ranking} \\
                            \cline{4-8}
&&\multicolumn{1}{l|}{Generalisation}      & 1  & 2 & 3 & 4 & 5\\ \hline \hline
& \multicolumn{2}{l|}{Written Language}     &      &      &      &    & \\
&\hspace*{.5cm}& language                   &  &  3 & 1  &  1     & \\ \hline
&& subject area/domain                      &  5.3   &  &  &       & \\ \hline
&& text type                                &  .3   &  .5 & 2 &  1  & \\ \hline
&& text length                              &  .3   &  1.5 & 1 & 1  & \\ \hline
& \multicolumn{2}{l|}{Spoken Language}      &    &  &  &       & \\
&& speaker-dependence                       &  4.25 &  & .5  &  & \\ \hline
&& spontaneity                              &  .25  &  2 &  .5 &  1 & \\ \hline
&& channel bandwidth                        &  .25  &  1.5 & 1 & 1 & \\ \hline
&& native/non-native speakers               &  .25  &  1.5 & 2 &  & \\ \hline
&& subject area/domain                      &   .5  &  & &   1   & \\ \hline
&\hspace*{.5cm}& language                   &   .5  &  &  &       & 1\\ \hline
\end{tabular}

\section*{References}

\begin{description}
\item[EAGLES 1994:] `Evaluation of Natural Language Processing Systems'
draft document EAG-EWG-IR.2, October 1994.

\item[Galliers \& Sparck Jones 1993:] `Evaluating Natural Language Processing 
Systems', Technical Report 291, Computer Laboratory, University of
Cambridge.

\item[Sparck Jones 1994:] `Towards better NLP System Evaluation', Proceedings
2nd ARPA Workshop on Human Language Technology
\end{description}

\end{document}